\documentclass[conference]{IEEEtran}
\IEEEoverridecommandlockouts

\usepackage{amsmath,amssymb,amsfonts}
\usepackage{booktabs}
\usepackage{multirow}
\usepackage{array}
\usepackage{makecell}
\usepackage{cellspace}
\usepackage{siunitx}
\usepackage{textcomp}

\usepackage{graphicx}
\graphicspath{{./Figures/}}
\usepackage{float}

\usepackage{xcolor}
\usepackage{colortbl}
\definecolor{lightgreen}{rgb}{0.9,1,0.9}

\usepackage{adjustbox}
\usepackage[normalem]{ulem}
\usepackage{xspace}

\usepackage[ruled,vlined]{algorithm2e}
\usepackage{algorithmic}

\usepackage{tikz}
\usetikzlibrary{positioning,backgrounds,shapes.geometric,calc,fit,arrows.meta}
\usepackage{pgfplots}
\usepackage{pgfplotstable}
\usepgfplotslibrary{polar}
\pgfplotsset{compat=1.18}

\usepackage{subcaption}

\usepackage{cite}
\usepackage[pagebackref,breaklinks,colorlinks]{hyperref}
\usepackage[capitalize]{cleveref}

\crefname{section}{Sec.}{Secs.}
\Crefname{section}{Section}{Sections}
\crefname{table}{Tab.}{Tabs.}
\Crefname{table}{Table}{Tables}
\crefname{figure}{Fig.}{Figs.}
\Crefname{figure}{Figure}{Figures}

\def\BibTeX{{\rm B\kern-.05em{\sc i\kern-.025em b}\kern-.08em
T\kern-.1667em\lower.7ex\hbox{E}\kern-.125emX}}

\tikzset{every mark/.append style={scale=0.8}}

\title{\LARGE \bf NPNet: A Non-Parametric Network with Adaptive Gaussian--Fourier Positional Encoding for 3D Classification and Segmentation}

\author{
	\parbox{\textwidth}{%
		\centering
		Mohammad Saeid$^{1}$, Amir Salarpour$^{2}$, Pedram MohajerAnsari$^{2}$, Mert D. Pes\'{e}$^{2}$%
	}%
	\thanks{$^{1}$Mohammad Saeid is with Sirjan University of Technology, Sirjan, Iran
		{\tt m.saeid@stu.sirjantech.ac.ir}}%
	\thanks{$^{2}$Amir Salarpour, Pedram MohajerAnsari, and Mert D. Pesé are with Clemson University, Clemson, SC, USA
		\{\texttt{asalarp}, \texttt{pmohaje}, \texttt{mpese}\}@clemson.edu}
}

\begin{document}
	
	\maketitle
	\thispagestyle{empty}
	\pagestyle{empty}
	
    \begin{abstract}

We present \textbf{NPNet}, a fully non-parametric approach for 3D point-cloud classification and part segmentation. NPNet contains no learned weights; instead, it builds point features using deterministic operators such as farthest point sampling, $k$-nearest neighbors, and pooling. Our key idea is an \emph{adaptive Gaussian--Fourier} positional encoding whose bandwidth and Gaussian--cosine mixing are chosen from the input geometry, helping the method remain stable across different scales and sampling densities. For segmentation, we additionally incorporate fixed-frequency Fourier features to provide global context alongside the adaptive encoding. Across ModelNet40/ModelNet-R, ScanObjectNN, and ShapeNetPart, NPNet achieves strong performance among non-parametric baselines, and it is particularly effective in few-shot settings on ModelNet40. NPNet also offers favorable memory use and inference time compared to prior non-parametric methods. Code is available at \href{https://github.com/m-saeid/NPNet}{GitHub: m-saeid/NPNet}.



\end{abstract}

    \section{Introduction}
\label{sec:intro}



Accurate and efficient 3D perception is central to intelligent vehicle systems. Point clouds captured by LiDAR and depth sensors support scene understanding, localization, and planning in autonomous driving~\cite{li2020deep,cui2021deep,abbasi2022lidar}. Similar needs also arise in robotics~\cite{duan2021robotics}, augmented reality~\cite{mahmood2020bim}, and digital-heritage mapping~\cite{melas2023pc2}. Beyond 3D perception, modern sensing pipelines in other application domains also face analogous calibration, prediction, and reliability challenges that motivate efficient and transferable models~\cite{rajoli2023thermal,saberian2024probabilistic,kokhahi2023glhad}. However, point clouds are irregularly sampled and often large, which makes it difficult to learn representations efficiently~\cite{lin2021pointacc}. This difficulty is amplified by real-time, on-vehicle constraints where compute and memory budgets are limited~\cite{hu2020randla,cao2025real}. In this setting, non-parametric architectures are a practical alternative. They rely on deterministic geometric operators rather than trainable weights, and they can transfer across scenes and sensors without retraining.



Parametric point-cloud networks such as PointNet/++~\cite{qi2017pointnet,qi2017pointnet++}, PointConv~\cite{wu2019pointconv}, KPConv~\cite{thomas2019kpconv}, DGCNN~\cite{wang2019dynamic}, PCT~\cite{guo2021pct}, and PointMLP~\cite{ma2022rethinking} achieve accuracy. However, they rely on millions of learned weights and typically require expensive training, which makes rapid adaptation difficult and can be limiting in few-shot settings~\cite{saeid2025enhancing,gu2024pointenet,sugiura2025pointode}. These limitations motivate non-parametric architectures that remove learned parameters while aiming to keep competitive performance.

Recent non-parametric models such as Point-NN~\cite{zhang2023parameter} and Point-GN~\cite{salarpour2024pointgn} are promising. However, their positional encodings are usually fixed, which can make them sensitive to changes in point density, object scale, and sampling distribution. As a result, performance often degrades when moving across datasets or when switching between tasks.


\begin{figure}[t]
    \centering
    \begin{tikzpicture}
    \begin{axis}[
        hide axis,
        xmin=0, xmax=1, ymin=0, ymax=1,
        legend columns=3,
        legend style={
            /tikz/every even column/.append style={column sep=1em},
            draw=none, fill=none,
            font=\scriptsize
        }
    ]
        \addlegendimage{area legend, fill=blue, opacity=0.5}
        \addlegendentry{NPNet}
        \addlegendimage{area legend, fill=red, opacity=0.5}
        \addlegendentry{Point-NN}
        \addlegendimage{area legend, fill=brown, opacity=0.5}
        \addlegendentry{Point-GN}
    \end{axis}
    \end{tikzpicture}

    \vspace{0.2em}

    \begin{subfigure}[t]{0.49\linewidth}
        \centering
        \begin{tikzpicture}
        \begin{polaraxis}[
            width=\linewidth,
            grid=both,
            xtick={0,90,180,270},
            xticklabels={
                \rotatebox{90}{Accuracy},
                Time,
                \rotatebox{90}{GFLOPs},
                Mem.
            },
            xticklabel style={font=\scriptsize, align=center},
            ymin=0, ymax=1,
            ytick={0,0.2,0.4,0.6,0.8,1},               
            yticklabel style={font=\tiny, opacity=1, scale=0.6}, 
            yticklabels={0,0.2,0.4,0.6,0.8,1},          
        ]
            \addplot+[mark=*, thick, fill=blue, fill opacity=0.05]
            coordinates {(0,1) (90,1) (180,1) (270,1) (360,1)};
            \addplot+[mark=*, thick, fill=red, fill opacity=0.04]
            coordinates {(0,0.95) (90,0.86) (180,0.77) (270,0.61) (360,0.95)};
            \addplot+[mark=*, color=brown, thick, fill=brown, fill opacity=0.03]
            coordinates {(0,0.99) (90,0.66) (180,1) (270,0.61) (360,0.99)};
        \end{polaraxis}
        \end{tikzpicture}
        \caption*{ModelNet40}
    \end{subfigure}
    \hfill
    \begin{subfigure}[t]{0.49\linewidth}
        \centering
        \begin{tikzpicture}
        \begin{polaraxis}[
            width=\linewidth,
            grid=both,
            xtick={0,90,180,270},
            xticklabels={
                \rotatebox{90}{mIoU},
                Time,
                \rotatebox{90}{GFLOPs},
                Mem.
            },
            xticklabel style={font=\scriptsize, align=center},
            ymin=0, ymax=1,
            ytick={0,0.2,0.4,0.6,0.8,1},               
            yticklabel style={font=\tiny, opacity=1, scale=0.6}, 
            yticklabels={0,0.2,0.4,0.6,0.8,1},          
        ]
            \addplot+[mark=*, thick, fill=blue, fill opacity=0.05]
            coordinates {(0,1) (90,1) (180,1) (270,1) (360,1)};
            \addplot+[mark=*, thick, fill=red, fill opacity=0.04]
            coordinates {(0,0.95) (90,0.33) (180,0.84) (270,0.57) (360,0.95)};
        \end{polaraxis}
        \end{tikzpicture}
        \caption*{ShapeNetPart}
    \end{subfigure}
    \caption{Radar comparison on ModelNet40 and ShapeNetPart. Metrics are normalized to $[0,1]$ per dataset (higher is better). GPU memory, GFLOPs, and inference time are inverted so larger values indicate better efficiency.}

    \label{fig:radar_combined}
    \vspace{-15pt}
\end{figure}
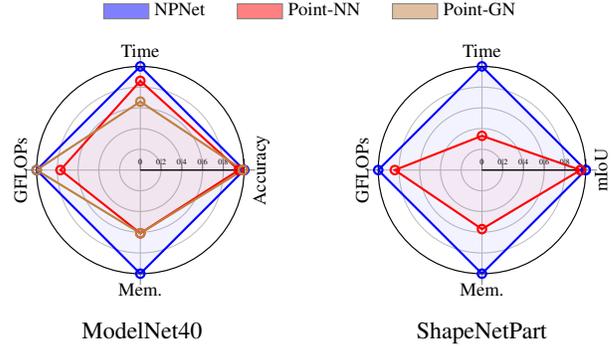

We introduce \textbf{NPNet}, a fully non-parametric framework for 3D point-cloud analysis. The core idea is an \emph{adaptive Gaussian--Fourier positional encoding} that chooses its bandwidth and Gaussian--cosine mixing from simple statistics of the input, which helps it stay stable across different scales and sampling densities. For segmentation, we also add fixed-frequency Fourier features to provide global context. This produces a hybrid encoding that combines local adaptivity with global structure without using any trainable weights. NPNet builds multi-scale features using only deterministic operators such as farthest point sampling, $k$-nearest-neighbor grouping, and pooling. For classification, we perform similarity-based inference, while a Fourier-augmented branch is used for segmentation. As shown in Fig.~\ref{fig:radar_combined}, NPNet improves both accuracy (or mIoU) and efficiency compared with prior zero-parameter methods on ModelNet40~\cite{Wu_2015_CVPR} and ShapeNetPart~\cite{chang2015shapenet}.

Across ModelNet40, ModelNet-R~\cite{saeid2025enhancing}, ScanObjectNN~\cite{uy2019revisiting}, and ShapeNetPart, NPNet achieves state-of-the-art performance among non-parametric methods and remains competitive with parametric networks. On ModelNet40 few-shot evaluation, it remains competitive, supporting the generalization benefits of the proposed adaptive, training-free encoding.

\begin{figure*}[t]
    \centering
    \includegraphics[width=1.7\columnwidth]{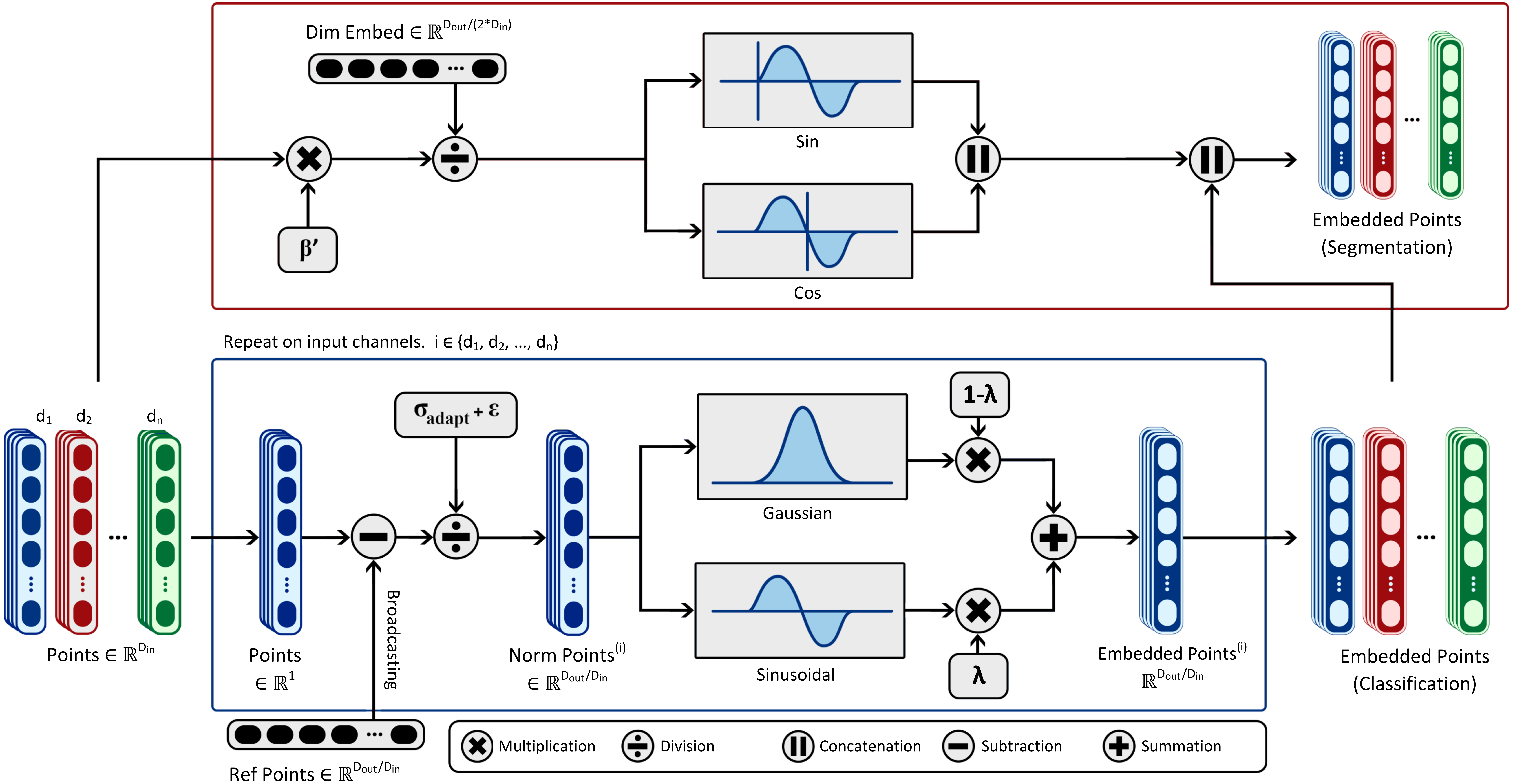}
    \caption{Adaptive Gaussian--Fourier positional encoding. The encoding adapts bandwidth $\sigma$ and mixing coefficient $\lambda$ from input geometry; an additional fixed-frequency Fourier branch provides global context for segmentation.}

    \label{fig:adaptive_encoding}
\end{figure*}

\vspace{13pt}
\noindent \textbf{Contributions.}
\begin{itemize}
  \item We propose NPNet, a fully non-parametric framework with a shared encoder for point-cloud classification and part segmentation.
  \item We introduce an adaptive Gaussian--Fourier positional encoding. It selects the bandwidth and the Gaussian--cosine mixing from input statistics, and it adds fixed-frequency Fourier features to provide global context for segmentation.
  \item We show that NPNet is competitive with, and often improves over, prior non-parametric baselines on standard benchmarks, including few-shot ModelNet40, while offering favorable memory use and inference time.
\end{itemize}

    \section{Related Work}
\label{sec:related_work}

\noindent \textbf{Representations.} Projection-based methods render point clouds into images or depth maps. This allows reuse of 2D CNNs, but it can introduce occlusion and discretization artifacts~\cite{ahn2022projection,jhaldiyal2023semantic,gopi2002fast}. Voxel and octree approaches enable 3D convolutions, yet they often incur high memory cost because of sparsity~\cite{zhou2018voxelnet,xu2021voxel,aljumaily2023point}. Point-based models operate directly on 3D coordinates and capture geometric detail, but they typically rely on large learned networks~\cite{qi2017pointnet,qi2017pointnet++,wang2019dynamic,thomas2019kpconv, peyghambarzadeh2020point}.

\noindent \textbf{Efficient and non-parametric models.} Hand-crafted pipelines and fixed transforms can reduce the need for learning, but they are often brittle in practice~\cite{zhang2020pointhop,kadam2022r, salarpour2025pointln}. More recent non-parametric networks such as Point-NN and Point-GN remove trainable weights by combining sampling, $k$-NN grouping, and fixed positional encodings~\cite{zhang2023parameter,salarpour2024pointgn}. However, these encodings use static bandwidths, which can make performance sensitive to changes in point density, object scale, and sampling distribution.

\noindent \textbf{Positional encoding.} Positional encodings play an important role in 3D learning~\cite{lu2022transformers}. Fixed trigonometric features and static Gaussian kernels can improve discrimination, but they effectively assume a single global setting. In contrast, \textbf{NPNet} uses an \emph{adaptive Gaussian--Fourier} encoding that selects the bandwidth and the Gaussian--cosine mixing from input statistics. For segmentation, we also add fixed-frequency Fourier terms to provide global context. This design keeps the simplicity and efficiency of non-parametric models while improving transfer across tasks and datasets.

    \section{Methodology}
\label{sec:methodogy}

\subsection{Overview of NPNet}
\label{sec:overview}

\textbf{NPNet} is a fully non-parametric framework for 3D point-cloud classification and part segmentation. It uses no trainable weights; instead, it builds features with deterministic geometric operators—farthest point sampling (FPS), $k$-nearest neighbors ($k$-NN), pooling—and an \emph{adaptive Gaussian--Fourier} positional encoding. For classification, a multi-stage encoder aggregates local neighborhoods into a single global descriptor and predicts labels by similarity matching to a memory bank of training shapes. For segmentation, an encoder--decoder produces pointwise descriptors and assigns part labels by matching to stored part prototypes. The whole pipeline is training-free: once the memory bank is built, inference reduces to encoding and nearest-prototype style matching.

\subsection{Adaptive Gaussian--Fourier Positional Encoding}
\label{sec:adaptive-encoding}

\begin{figure*}[t]
    \centering
    \includegraphics[width=1.8\columnwidth]{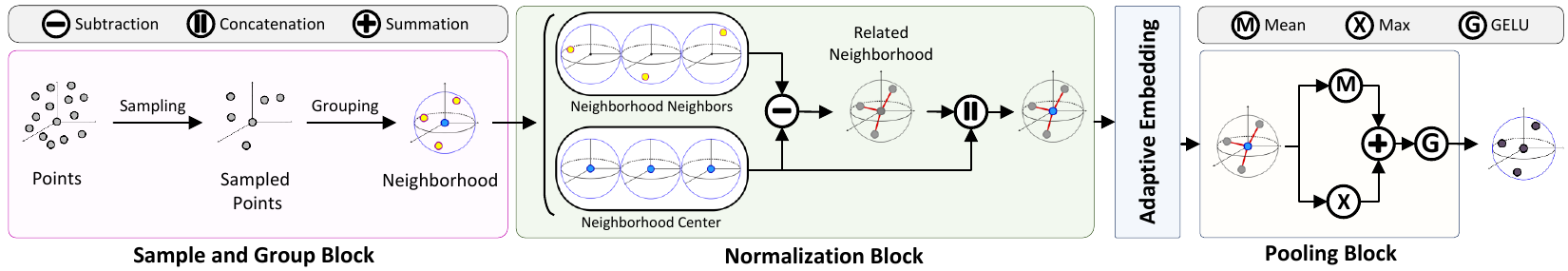}
    \caption{Stage block used in NPNet. FPS selects centroids, $k$-NN groups local neighborhoods, positional encoding modulates features, and mean/max pooling produces a stage descriptor; concatenating stages forms a multi-scale representation.}

    \label{fig:stage}
\end{figure*}

Given coordinates $X\in\mathbb{R}^{N\times 3}$, we build positional features with two components: (i) an \emph{adaptive} channel that blends Gaussian RBF and cosine responses using simple input statistics, and (ii) a \emph{Fourier} channel (segmentation only) that adds fixed multi-frequency context. Fig.~\ref{fig:adaptive_encoding} illustrates the encoding itself and Fig.~\ref{fig:stage} shows where the encoding is injected within each stage.


\paragraph{Adaptive channel.}
Let $\sigma_{\mathrm{g}}=\tfrac{1}{3}\sum_{i=1}^{3}\mathrm{Std}(X_{:,i})$ denote a global dispersion statistic. We set the bandwidth
$\sigma_{\mathrm{a}}=\sigma_0(1+\sigma_{\mathrm{g}})$ and the blend
$\lambda=\mathrm{sigmoid}((\sigma_{\mathrm{g}}-\tau)\kappa)$.
For anchors $\{v_m\}_{m=1}^{M}$ (fixed reference locations) and a scalar coordinate $x$,
\[
\begin{aligned}
\phi_{\mathrm{RBF}}(x,v_m) &= \exp\!\left(-\tfrac12\left(\tfrac{x-v_m}{\sigma_{\rm a}+\epsilon}\right)^2\right),\\
\phi_{\cos}(x,v_m) &= \cos\!\left(\tfrac{x-v_m}{\sigma_{\rm a}+\epsilon}\right),
\end{aligned}
\]
and the adaptive response is
\[
\phi_{\mathrm{adaptive}}(x)=\lambda\,\phi_{\mathrm{RBF}}(x)+(1-\lambda)\,\phi_{\cos}(x).
\]
Concatenation over anchors and coordinate channels yields $H_{\mathrm{adaptive}}\in\mathbb{R}^{N\times d'}$.

\paragraph{Fourier channel (segmentation).}
With frequencies $\omega_j=\alpha^{j/L}$ for $j{=}1\ldots L$ and scale $\beta$,

\[
\begin{aligned}
\phi_{\mathrm{Fourier}}(x)=[\sin(\beta x/\omega_1),\cos(\beta x/\omega_1),\\\ldots,\sin(\beta x/\omega_L),\cos(\beta x/\omega_L)].
\end{aligned}
\]

\paragraph{Hybrid code and neighborhood modulation.}
Classification uses only $H_{\mathrm{adaptive}}$; segmentation concatenates
\[
H_{\mathrm{pos}}=\big[\,H_{\mathrm{Fourier}} \,\|\, H_{\mathrm{adaptive}}\,\big].
\]
Within each $k$-NN neighborhood $\mathcal{N}$ (using relative coordinates), features are modulated by
\begin{equation}
\tilde{H}_{\mathcal{N}}=(H_{\mathcal{N}}+H_{\mathrm{pos}})\odot H_{\mathrm{pos}},
\label{eq:modulate}
\end{equation}
where $\odot$ denotes element-wise multiplication.

\begin{figure}[!b]
    \centering
    \includegraphics[width=1\columnwidth]{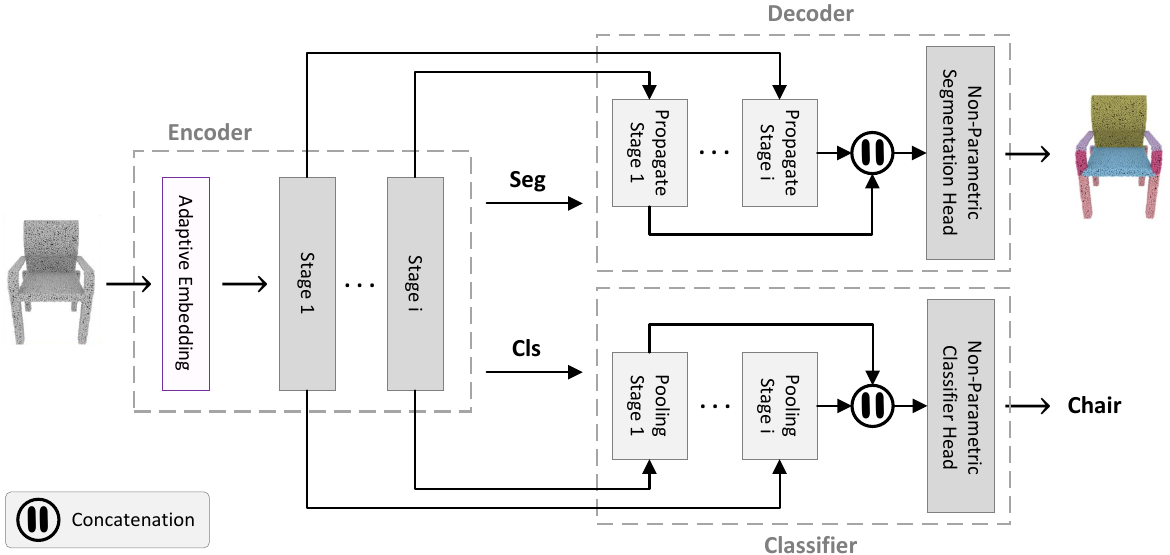}
    \caption{NPNet architecture for classification and part segmentation. The model contains no trainable weights; inference is performed by similarity matching to stored shape descriptors (classification) or part prototypes (segmentation).}
 
    \label{fig:arch}
\end{figure}

\subsection{NPNet Classification Architecture}
\label{sec:npnet-cls}

A $T$-stage encoder builds a feature pyramid by repeating FPS to $N_t$ centroids, $k$-NN grouping, adaptive modulation (Eq.~\ref{eq:modulate}), and mean+max pooling (Fig.~\ref{fig:stage}). The full architecture is shown in Fig.~\ref{fig:arch}. Let $F^{(t)}_j$ be the pooled feature of centroid $j$ in stage $t$. We form the global descriptor by concatenating stage summaries:
\[
F^{\mathrm{enc}}=\big\Vert_{t=1}^{T}\big[\,\max\nolimits_j F^{(t)}_j \,\|\, \mathrm{mean}_j F^{(t)}_j\,\big]\in\mathbb{R}^{D},
\]
with normalization/nonlinearities and \emph{no learned layers}. This descriptor is used for similarity-based inference (Sec.~\ref{sec:npnet-infer}).

\subsection{NPNet Segmentation Architecture}
\label{sec:seg-arch}

The segmentation encoder mirrors the classification backbone but uses the hybrid code $H_{\mathrm{pos}}$. A non-parametric decoder propagates coarse features to finer resolutions via inverse-distance weighting (IDW). For point $x_i$ at level $t$ with neighbors in level $t{+}1$,
\begin{equation}
\hat{h}^{(t)}_i=\sum_{j\in\mathcal{N}(x_i,X^{(t+1)})}
\frac{\|x_i-x^{(t+1)}_j\|_2^{-1}}{\sum_k \|x_i-x^{(t+1)}_k\|_2^{-1}}\;h^{(t+1)}_j .
\label{eq:idw}
\end{equation}
The interpolated features $\hat{H}^{(t)}$ are concatenated with encoder features $H^{(t)}$ and propagated to the next finer level until reaching the input resolution, yielding per-point descriptors.

\subsection{Non-Parametric Training and Inference}
\label{sec:npnet-infer}
NPNet follows a simple routine: encode the training set once to build a memory bank, then predict by similarity—no gradient updates.

\paragraph{Classification.}
For training shapes $(X_i,y_i)$, compute normalized descriptors
$f_i=\mathcal{E}_{\mathrm{cls}}(X_i)/\|\mathcal{E}_{\mathrm{cls}}(X_i)\|_2$ and store
$F=[f_1,\ldots,f_{M_{\mathrm{tr}}}]\in\mathbb{R}^{D\times M_{\mathrm{tr}}}$ with one-hot labels $Y\in\{0,1\}^{M_{\mathrm{tr}}\times C}$.
For a test feature $f$, similarities $s=f^\top F$ yield weights
$w=\mathrm{softmax}(\gamma s)$ (temperature $\gamma$). The prediction is
$\hat{z}=w^\top Y$ and $\hat{y}=\arg\max_c \hat{z}_c$.

\paragraph{Segmentation.}
For each training shape, extract per-point features $F_i=\mathcal{E}_{\mathrm{seg}}(X_i)$ and form part prototypes $\mu_{i,p}$ by averaging features over points with label $p$.
Store prototypes per shape category $s$ as $(F_s,Y_s)$.
For a test point feature $f_j$ (category $s$), compute $s_j=f_j^\top F_s$,
$w_j=\mathrm{softmax}(\gamma s_j)$, $\hat{z}_j=w_j^\top Y_s$; assign
$\hat{y}_j=\arg\max_p \hat{z}_{j,p}$.

    \section{Experiments}
\label{sec:experiments}

\begin{table}[!t]
\scriptsize
\centering
\caption{ModelNet40 shape classification. Accuracy (\%), trainable parameters (M), and GFLOPs (G).}
\begin{tabular}{lccc}
\hline
\textbf{Method} & \textbf{Acc. (\%)} & \textbf{Param (M)} & \textbf{GFLOPs (G)} \\ \hline
\multicolumn{3}{l}{\textit{Parametric Methods}} \\
PointNet~\cite{qi2017pointnet} & 89.2 & 3.5 & 0.4 \\
PointNet++~\cite{qi2017pointnet++} & 90.7 & 1.5 & 0.8 \\
PointCNN~\cite{li2018pointcnn} & 92.2 & 0.6 & - \\
PointConv~\cite{wu2019pointconv} & 92.5 & 18.6 & - \\
OctFormer~\cite{wang2023octformer} & 92.7 & 4 & 31.3 \\
KPConv~\cite{thomas2019kpconv} & 92.9 & 15.2 & - \\
DGCNN~\cite{wang2019dynamic} & 92.9 & 1.8 & 2.7 \\
PCT~\cite{guo2021pct} & 93.2 & 2.9 & 2.3 \\
PointNext-S~\cite{qian2022pointnext} & 93.2 & 1.4 & - \\
GBNet~\cite{qiu2021geometric} & 93.8 & 8.4 & - \\
CurveNet~\cite{xiang2021walk} & 93.8 & 2.1 & 0.3 \\
PointMLP~\cite{ma2022rethinking} & \textbf{94.1} & 13.2 & 15.7 \\
\hline
\multicolumn{3}{l}{\textit{Non-parametric Methods}} \\
Point-NN~\cite{zhang2023parameter} & 81.8 & 0.0 & 0.0 \\ 
Point-GN~\cite{salarpour2024pointgn} & 85.3 & 0.0 & 0.0 \\
\rowcolor{lightgreen} NPNet (ours) & \textbf{85.45} & 0.0 & 0.0 \\
\end{tabular}

\label{tab:modelnet40}
\end{table}

We evaluate NPNet on standard benchmarks for shape classification, few-shot classification, and part segmentation. We summarize datasets and metrics, describe the training-free setup and hyperparameters, then report results and efficiency. Finally, we ablate the adaptive Gaussian--Fourier encoding and key design choices.

\subsection{Datasets and Evaluation Metrics}
\noindent\textbf{ModelNet40:} 40 CAD categories (9{,}843 train / 2{,}468 test). Metric: overall accuracy (Tab.~\ref{tab:modelnet40}).\\
\textbf{ModelNet-R:} relabeled/cleaned ModelNet40 variant. Metric: overall accuracy (Tab.~\ref{tab:modelnetr}).\\
\textbf{ScanObjectNN:} real-world RGB-D scans with clutter, occlusion, and noise. Metric: accuracy on OBJ-BG, OBJ-ONLY, and PB-T50-RS (Tab.~\ref{tab:scanobject}).\\
\textbf{Few-shot ModelNet40:} $N$-way/$K$-shot episodic classification. Metric: mean accuracy (Tab.~\ref{tab:fewshot}).\\
\textbf{ShapeNetPart:} 16 categories with 50 part labels. Metric: instance mIoU (Tab.~\ref{tab:shapenet}).

\subsection{Implementation Details and NPNet Hyperparameters}
\label{sec:impl_details}

NPNet is training-free: one forward pass over the training set builds the memory banks (shape descriptors for classification; part prototypes for segmentation). Unless stated otherwise, we sample $N{=}1{,}024$ points per shape and disable augmentation to isolate the effect of the adaptive encoding.

\noindent\textbf{Batch sizes and hardware.}
ModelNet40 and few-shot use batch size 10, ScanObjectNN 16, and ShapeNetPart 1 (per-shape). All runs use a single NVIDIA GeForce RTX~3090 (24\,GB, CUDA~11.5).

\begin{table}[!t] 
\scriptsize
\centering
\caption{ModelNet-R shape classification. Accuracy (\%), trainable parameters (M), and GFLOPs (G) where available.}

\begin{tabular}{lccc}
\hline
\textbf{Method} & \textbf{Acc. (\%)} & \textbf{Param (M)} & \textbf{GFLOPs (G)} \\ \hline
\multicolumn{3}{l}{\textit{Parametric Methods}} \\
PointNet \cite{qi2017pointnet} & 91.4 & 3.5 & 0.4 \\
PointNet++ (SSG) \cite{qi2017pointnet++} & 94.0 & 1.5 & 0.8 \\
PointNet++ (MSG) \cite{qi2017pointnet++} & 94.0 & 1.7 & 4.0 \\
DGCNN \cite{wang2019dynamic} & 94.0 & 1.8 & 2.7 \\
CurveNet \cite{xiang2021walk} & 94.1 & 2.1 & 0.3 \\
Point-SkipNet \cite{saeid2025enhancing} & 94.3 & 1.5 & - \\
PointMLP \cite{ma2022rethinking} & \textbf{95.3} & 13.2 & 15.7 \\
\hline

\multicolumn{3}{l}{\textit{Non-parametric Methods}} \\
Point-NN \cite{zhang2023parameter} & 84.75 & \textbf{0.0} & \textbf{0.0} \\

\rowcolor{lightgreen} NPNet (ours) & \textbf{85.65} & \textbf{0.0} & \textbf{0.0} \\
\end{tabular}

\label{tab:modelnetr}
\end{table}

\noindent\textbf{Baseline hyperparameters.}
\textbf{ModelNet40/ModelNet-R/few-shot:} $d{=}35$, $k{=}110$, stages$=4$, $N{=}1{,}024$.\\
\textbf{ScanObjectNN:} $d{=}27$, $k{=}120$, stages$=4$, $N{=}1{,}024$.\\
\textbf{ShapeNetPart:} $d{=}144$, $k{=}70$, stages$=2$, $N{=}1{,}024$.\\
These settings follow Sec.~\ref{sec:ablation} and balance accuracy against runtime and memory.

\subsection{Classification Results}
On \textbf{ModelNet40} and \textbf{ModelNet-R} (Tabs.~\ref{tab:modelnet40},~\ref{tab:modelnetr}), NPNet achieves \textbf{85.45\%} and \textbf{85.65\%} accuracy with \textbf{0.0M} parameters, outperforming prior non-parametric methods.
On \textbf{ScanObjectNN} (Tab.~\ref{tab:scanobject}), NPNet reaches \textbf{86.1\%}/\textbf{86.1\%}/\textbf{84.9\%} on OBJ-BG/OBJ-ONLY/PB-T50-RS, leading non-parametric baselines on OBJ-BG and OBJ-ONLY while remaining competitive on PB-T50-RS. Recent parametric models score higher but rely on millions of trainable weights; NPNet offers a strong training-free alternative.

\subsection{Few-Shot Classification}
Few-shot evaluation matches NPNet’s inference: each episode is solved by building descriptors and matching prototypes, with no fine-tuning. Averaged over 10 runs on ModelNet40, NPNet attains \textbf{92.0\%}/\textbf{93.2\%} (5-way 10-/20-shot) and \textbf{82.5\%}/\textbf{87.6\%} (10-way 10-/20-shot). Tab.~\ref{tab:fewshot} shows NPNet is best among the listed methods across all settings.

\subsection{Segmentation Results}
Tab.~\ref{tab:shapenet} reports ShapeNetPart results. NPNet achieves \textbf{73.56\%} instance mIoU using the hybrid Gaussian--Fourier encoding. Compared with Point-NN (70.4\%), the gain suggests that adding fixed-frequency Fourier terms alongside the adaptive channel helps capture global structure and repeated patterns that are useful for part boundaries. We use the baseline hyperparameters from Sec.~\ref{sec:impl_details} ($d{=}144$, $k{=}70$, stages$=2$).

\begin{table}[b]
\scriptsize
\centering
\caption{ScanObjectNN classification. Accuracy (\%) on OBJ-BG, OBJ-ONLY, and PB-T50-RS, with trainable parameters (M).}
\begin{tabular}{lcccc}
\toprule
\textbf{Method} & \textbf{OBJ} & \textbf{OBJ-} & \textbf{PB-T50} & \textbf{Param} \\
\textbf{} & \textbf{-BG} & \textbf{ONLY} & \textbf{-RS} & \textbf{(M)} \\
\midrule
\multicolumn{4}{l}{\textit{Parametric Methods}} \\
PointNet \cite{qi2017pointnet} & 73.3 & 79.2 & 68.2 & 3.5  \\
PointNet++ \cite{qi2017pointnet++} & 82.3 & 84.3 & 77.9 & 1.5  \\
DGCNN \cite{wang2019dynamic} & 82.8 & \textbf{86.2} & 78.1 & 1.8  \\
PointCNN \cite{li2018pointcnn} & \textbf{86.1} & 85.5 & 78.5 & - \\
GBNet \cite{qiu2021geometric} & - & - & 80.5 & 8.4  \\
PointMLP \cite{ma2022rethinking} & - & - & 85.4 & 12.6  \\
PointNeXt-S \cite{qian2022pointnext} & - & - & 87.7 & 1.5  \\
PointMetaBase-S \cite{lin2023meta} & - & - & \textbf{87.9} & 0.6  \\
\midrule
\multicolumn{3}{l}{\textit{Non-parametric Methods}} \\
Point-NN \cite{zhang2023parameter} & 71.1 & 74.9 & 64.9 & 0.0  \\
Point-GN \cite{salarpour2024pointgn} & 85.2 & 86.0 & \textbf{86.4} & 0.0  \\
\rowcolor{lightgreen} NPNet (ours) & \textbf{86.1} & \textbf{86.1} & 84.9 & 0.0  \\
\end{tabular}

\label{tab:scanobject}
\end{table}

\subsection{Efficiency and Deployment Analysis}
\label{sec:efficiency and deployment analysis}
NPNet’s cost profile is driven by neighborhood aggregation and positional encoding rather than learned layers: (i) \textbf{no training} (a single pass builds the banks), (ii) \textbf{inference} dominated by encoding, grouping, and lightweight similarity, and (iii) \textbf{memory/runtime} controlled mainly by $d$ and $k$. On \textbf{ModelNet40} (Tab.~\ref{tab:efficiency}), NPNet uses \textbf{0.0021 GFLOPs}, \textbf{99.1\,MB} GPU memory, and \textbf{3.86\,ms/sample}; on \textbf{ShapeNetPart} it uses \textbf{0.0045 GFLOPs}, \textbf{256.4\,MB}, and \textbf{5.63\,ms/sample}---faster and leaner than Point-NN/Point-GN.

\noindent\textbf{One-off bank construction and scaling.}
NPNet incurs a one-time preprocessing cost to build the memory banks: for classification, storing one descriptor per training shape; for segmentation, storing part prototypes per category. This cost is linear in the number of training shapes and is typically amortized over many test queries. The bank memory footprint scales with descriptor dimension, i.e., $\mathcal{O}(M_{\text{train}}D)$ for classification and $\mathcal{O}(M_{\text{proto}}D)$ for segmentation, where $M_{\text{proto}}$ depends on the number of shapes and parts per category. When memory is constrained, descriptor compression (e.g., FP16 storage) or prototype subsampling/clustering can reduce the footprint with minimal impact. For very large banks, approximate nearest-neighbor search can further reduce query time while keeping the training-free pipeline unchanged.

\subsection{Discussion: Complexity, Scope, and Reproducibility}
\label{sec:discussion}

\begin{table}[!t] 
    \scriptsize
    \centering
    \renewcommand{\arraystretch}{1.2}
    \caption{Few-shot classification on ModelNet40. Mean accuracy (\%) over 10 runs for 5-way and 10-way tasks; baseline results are taken from the cited reference.}

    \begin{tabular}{lcccc}
        \hline
        \textbf{Method} & \multicolumn{2}{c}{\textbf{5-way}} & \multicolumn{2}{c}{\textbf{10-way}} \\
        & \textbf{10-shot} & \textbf{20-shot} & \textbf{10-shot} & \textbf{20-shot} \\
        \hline
        \multicolumn{5}{l}{\textit{Parametric Methods}} \\
        DGCNN~\cite{wang2019dynamic} & 31.6 & 40.8 & 19.9 & 16.9 \\
        FoldingNet~\cite{yang2018foldingnet} & 33.4 & 35.8 & 18.6 & 15.4 \\
        PointNet++~\cite{qi2017pointnet++} & 38.5 & 42.4 & 23.0 & 18.8 \\
        PointNet~\cite{qi2017pointnet} & 52.0 & 57.8 & 46.6 & 35.2 \\
        3D-GAN~\cite{wu2016learning} & 55.8 & 65.8 & 40.3 & 48.4 \\
        PointCNN~\cite{li2018pointcnn} & 65.4 & 68.6 & 46.6 & 50.0 \\
        \hline
        \multicolumn{5}{l}{\textit{Non-parametric Methods}} \\
        Point-NN~\cite{zhang2023parameter} & 88.8 & 90.9 & 79.9 & 84.9 \\
        Point-GN~\cite{salarpour2024pointgn} & 90.7 & 90.9 & 81.6 & 86.4 \\ 

        \rowcolor{lightgreen} NPNet (ours) & \textbf{92.0} & \textbf{93.2} & \textbf{82.5} & \textbf{87.6} \\

    \end{tabular}

    \label{tab:fewshot}
\end{table}

\noindent \textbf{Complexity and scaling.}
Let $N$ be the number of input points, $T$ the number of stages, $k$ the neighborhood size, and $d$ the embedding dimension. At stage $t$, FPS selects $N_t$ centroids, and each centroid aggregates a $k$-NN neighborhood with modulation (Sec.~\ref{sec:adaptive-encoding}) and mean--max pooling. The per-stage work scales as $\mathcal{O}(N_t k d)$ and memory as $\mathcal{O}(N_t d + k d)$, so the encoder cost is $\sum_{t=1}^{T}\mathcal{O}(N_t k d)$. Segmentation adds IDW upsampling with the same order (Eq.~\ref{eq:idw}). In practice, $k$ and $d$ are the main levers: larger $d$ improves accuracy, larger $k$ improves context but increases cost, and deeper hierarchies help classification more than segmentation.

\begin{table}[!t] 
\scriptsize
\centering
\caption{ShapeNetPart part segmentation. Instance mIoU (\%), input points, trainable parameters (M), GFLOPs (G), and training time (when reported).}

\begin{tabular}{lccccc}
\hline

\textbf{Method} & \textbf{Inputs} & \textbf{Inst.} & \textbf{Param} & \textbf{GFLOPs} & \textbf{Train} \\

\textbf{} & \textbf{} & \textbf{mIoU} & \textbf{(M)} & \textbf{(G)} & \textbf{Time}

\\ \hline
\multicolumn{3}{l}{\textit{Parametric Methods}} \\
PointNet \cite{qi2017pointnet}      & 2k & 83.7 & 8.3 & 5.8 & - \\
PointNet++ \cite{qi2017pointnet++}    & 2k & 85.1 & 1.8 & 1.1 & 26.5 h \\
DGCNN \cite{wang2019dynamic}         & 2k & 85.2 & 1.4 & 4.9 & - \\
APES(local) \cite{wu2023attention}  & 2k & 85.6 & 2.0 & 18.37 & - \\
APES(global) \cite{wu2023attention} & 2k & 85.8 & 2.0 & 15.56 & - \\
PAConv \cite{xu2021paconv}        & 2k & 86.0 & - & - & - \\
PointMLP \cite{ma2022rethinking}      & 2k & 86.1 & 16.8 & 6.2 & 47.1 h \\
CurveNet \cite{xiang2021walk}      & 2k & \textbf{86.6} & 5.5  & 2.5 & 56.9 h \\
\hline
\multicolumn{5}{l}{\textit{Non-parametric Methods}} \\
Point-NN \cite{zhang2023parameter}      & 1k & 70.4 & 0.0 & 0.0 & 0.0  \\ 
\rowcolor{lightgreen} NPNet (ours) & 1k & \textbf{73.56} & 0.0 & 0.0 & 0.0 \\
\end{tabular}

\label{tab:shapenet}
\end{table}
\begin{table}[!h]
\scriptsize
\centering
\caption{Efficiency of non-parametric methods. GFLOPs (G), peak GPU memory (MB), and inference time (ms/sample) on ModelNet40 and ShapeNetPart using 1{,}024 points; measured on an RTX 3090.}

\begin{tabular}{lccccc}
\toprule
\textbf{Method} & \textbf{Dataset} & \textbf{GFLOPs} & \textbf{Mem.} & \textbf{Inference}& \textbf{Num} \\
\textbf{} & \textbf{} & \textbf{(G)} & \textbf{(MB)} & \textbf{(ms/sample)} & \textbf{Points} \\
\midrule
Point-NN \cite{zhang2023parameter} & ModelNet40   & 0.0027 & 161.0  & 4.44  & 1024 \\
Point-GN \cite{salarpour2024pointgn} & ModelNet40   & \textbf{0.0021} & 161.0  & 5.80 & 1024 \\
\rowcolor{lightgreen} NPNet (ours)   & ModelNet40   & \textbf{0.0021} & \textbf{99.1}   & \textbf{3.86}  & 1024   \\
\midrule
Point-NN \cite{zhang2023parameter} & ShapeNetPart   & 0.0054 & 442.9  & 16.83 & 1024 \\
\rowcolor{lightgreen} NPNet (ours)   & ShapeNetPart   & \textbf{0.0045} & \textbf{256.4}  & \textbf{5.63}  & 1024  \\
\end{tabular}

\label{tab:efficiency}
\end{table}

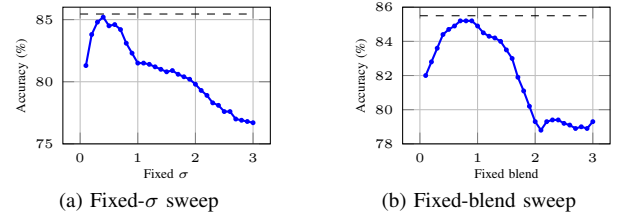
\begin{figure}[b]
  \centering
  \captionsetup[subfigure]{justification=centering, font=footnotesize, skip=1pt}

  \begin{subfigure}[t]{0.49\columnwidth}
    \centering
    \begin{tikzpicture}
      \begin{axis}[
        width=\linewidth,
        height=3.4cm,
        xlabel={Fixed $\sigma$},
        ylabel={Accuracy (\%)},
        xlabel near ticks,
        xlabel style={yshift=4pt},
        ylabel near ticks,
        ylabel style={yshift=-4pt},
        ymin=75, ymax=86,
        grid=major,
        tick label style={font=\tiny},
        label style={font=\tiny},
      ]
      \addplot[mark=*, mark size=0.65pt, thick, blue] coordinates {
        (0.1,81.3) (0.2,83.8) (0.3,84.8)
        (0.4,85.2) (0.5,84.5) (0.6,84.6) (0.7,84.2)
        (0.8,83.1) (0.9,82.3) (1,81.5) (1.1,81.5)
        (1.2,81.4) (1.3,81.2) (1.4,81.0) (1.5,80.8)
        (1.6,80.9) (1.7,80.6) (1.8,80.4) (1.9,80.2)
        (2,79.8) (2.1,79.3) (2.2,78.9) (2.3,78.3)
        (2.4,78.1) (2.5,77.6) (2.6,77.6) (2.7,77.0)
        (2.8,76.9) (2.9,76.8) (3,76.7)
      };
      \addplot[dashed, thin] coordinates {(0,85.45) (3,85.45)};
      \end{axis}
    \end{tikzpicture}
    \caption{Fixed-$\sigma$ sweep}
    \label{fig:fixsigma}
  \end{subfigure}
  \hfill
  \begin{subfigure}[t]{0.49\columnwidth}
    \centering
    \begin{tikzpicture}
      \begin{axis}[
        width=\linewidth,
        height=3.4cm,
        xlabel={Fixed blend},
        ylabel={Accuracy (\%)},
        xlabel near ticks,
        xlabel style={yshift=4pt},
        ylabel near ticks,
        ylabel style={yshift=-4pt},
        ymin=78, ymax=86,
        grid=major,
        tick label style={font=\tiny},
        label style={font=\tiny},
      ]
      \addplot[mark=*, mark size=0.65pt, thick, blue] coordinates {
        (0.1,82.0) (0.2,82.8) (0.3,83.6)
        (0.4,84.4) (0.5,84.7) (0.6,84.9) (0.7,85.2)
        (0.8,85.2) (0.9,85.2) (1,84.9) (1.1,84.5)
        (1.2,84.3) (1.3,84.2) (1.4,84.0) (1.5,83.5)
        (1.6,83.0) (1.7,81.9) (1.8,81.1) (1.9,80.2)
        (2,79.3) (2.1,78.8) (2.2,79.3) (2.3,79.4)
        (2.4,79.4) (2.5,79.2) (2.6,79.1) (2.7,78.9)
        (2.8,79.0) (2.9,78.9) (3,79.3)
      };
      \addplot[dashed, thin] coordinates {(0,85.5) (3,85.5)};
      \end{axis}
    \end{tikzpicture}
    \caption{Fixed-blend sweep}
    \label{fig:fixblend}
  \end{subfigure}
  \caption{Effect of disabling adaptivity on ModelNet40. Accuracy when (a) $\sigma$ is fixed and (b) the Gaussian--cosine mixing ratio is fixed in the positional encoding.}

  \label{fig:fix}
\end{figure}

\noindent \textbf{Assumptions and limitations.}
(1) \textit{Rotation equivariance:} the encoding is not rotation-equivariant; canonical alignment or test-time rotation averaging can help.
(2) \textit{Category for segmentation:} following ShapeNetPart protocol, the object category is known at test time; removing this assumption is future work.
(3) \textit{Memory-bank size:} the bank grows with the number of training shapes/prototypes; clustering or subsampling can cap memory with modest impact.
(4) \textit{$k$-NN backend:} very sparse or very large $N$ may benefit from approximate neighbors; NPNet is compatible with GPU ANN backends.

\begin{figure*}[t]
  \centering
  \resizebox{\textwidth}{!}{%
    \setlength{\tabcolsep}{2pt}
    \begin{tabular}{c *{7}{c}}

      \raisebox{0.4\height}{\rotatebox{90}{\tiny \textbf{DGCNN}}} &
      \includegraphics[width=0.11\textwidth]{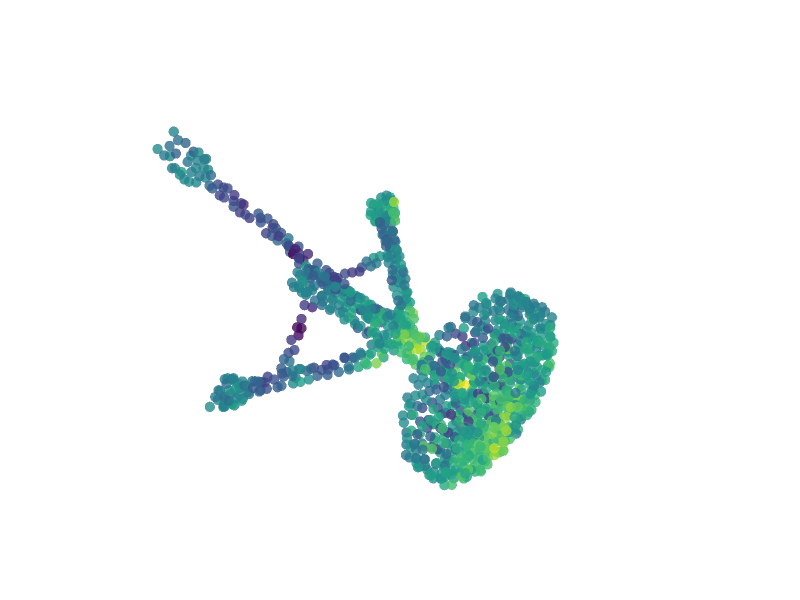} &
      \includegraphics[width=0.11\textwidth]{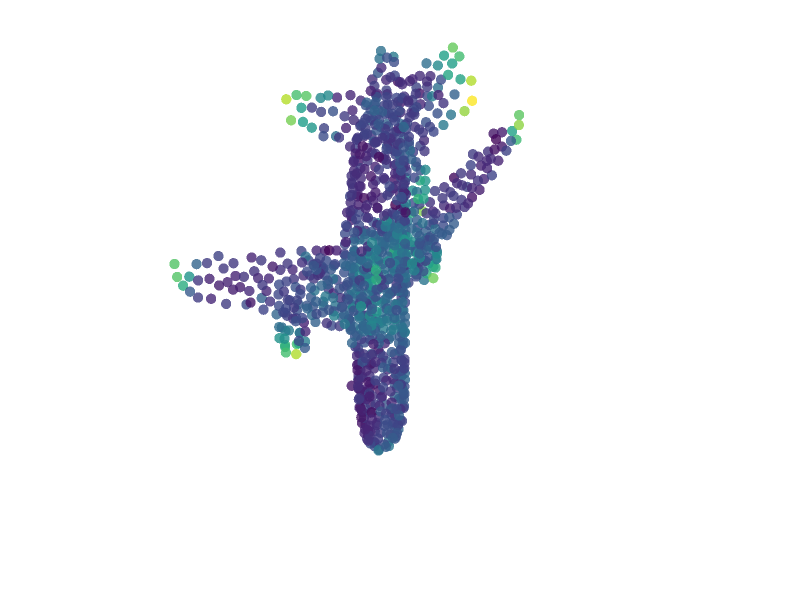} &
      \includegraphics[width=0.11\textwidth]{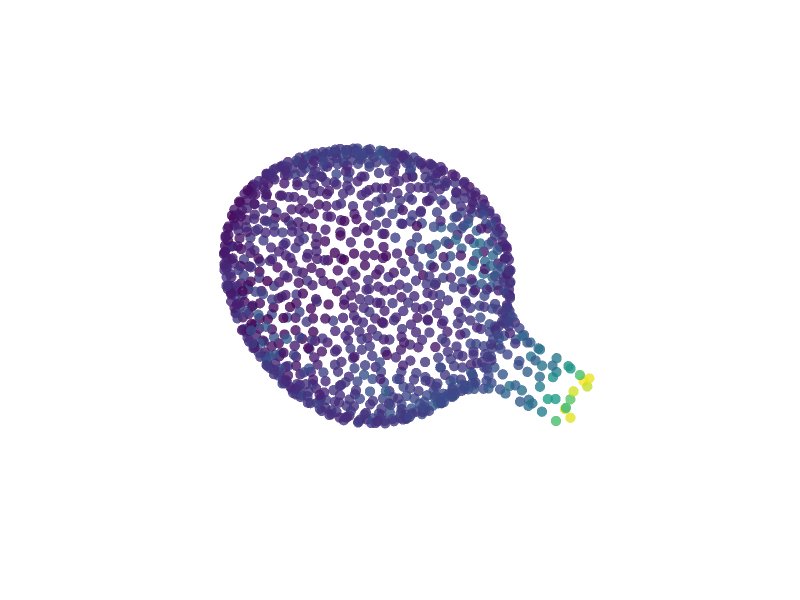} &
      \includegraphics[width=0.11\textwidth]{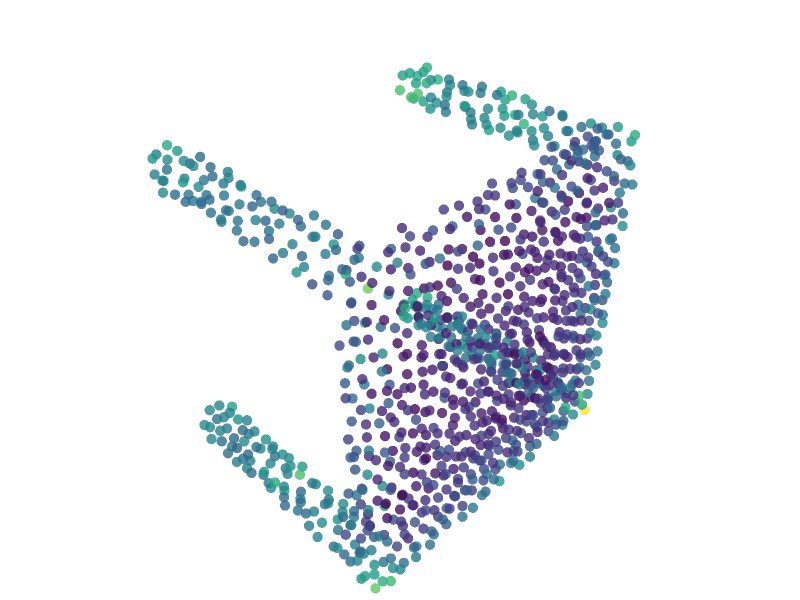} &
      \includegraphics[width=0.11\textwidth]{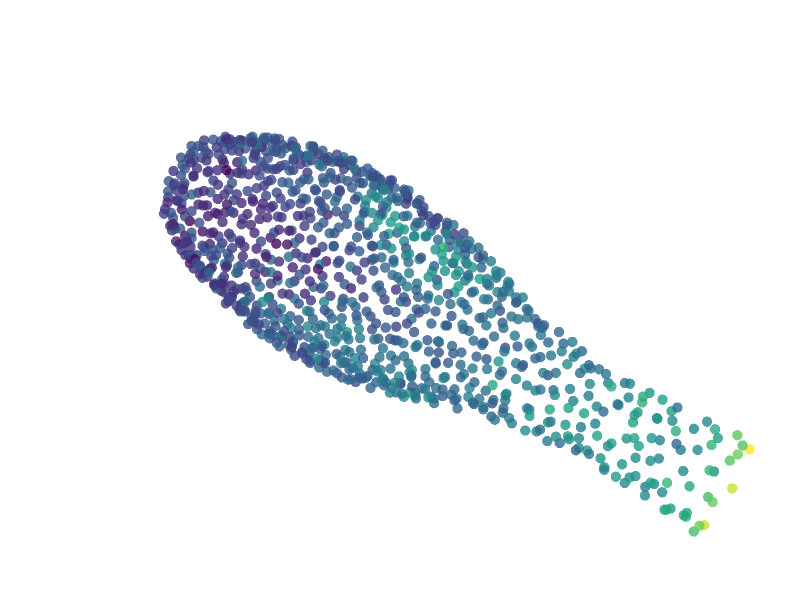} &
      \includegraphics[width=0.11\textwidth]{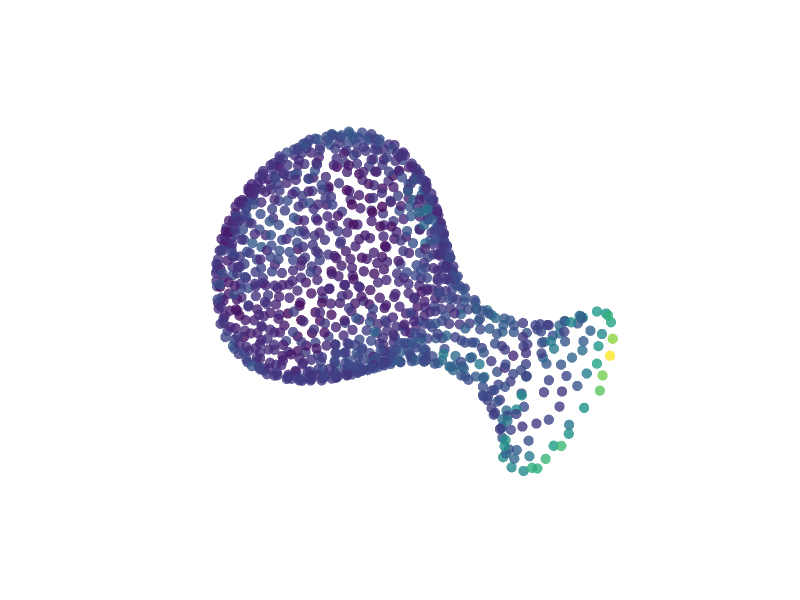} &
      \includegraphics[width=0.11\textwidth]{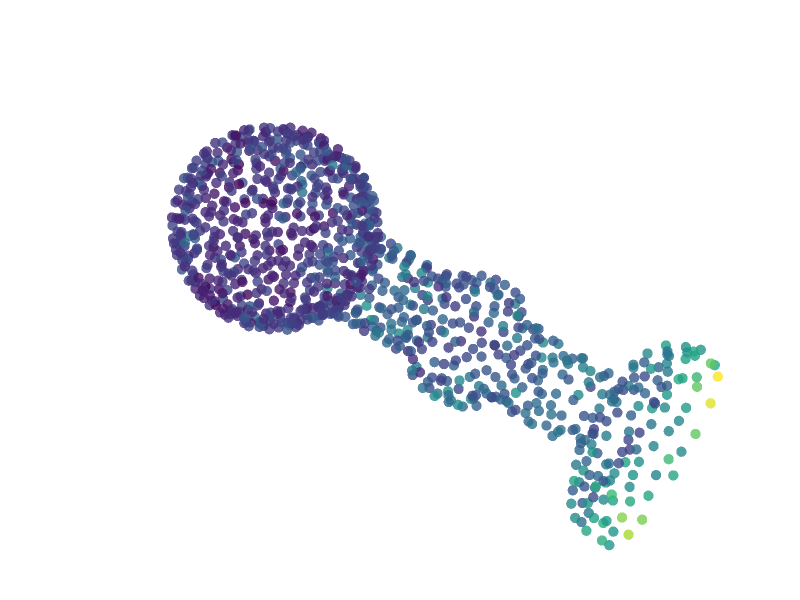} \\
      
      \raisebox{0.4\height}{\rotatebox{90}{\tiny \textbf{Adaptive16}}} &
      \includegraphics[width=0.11\textwidth]{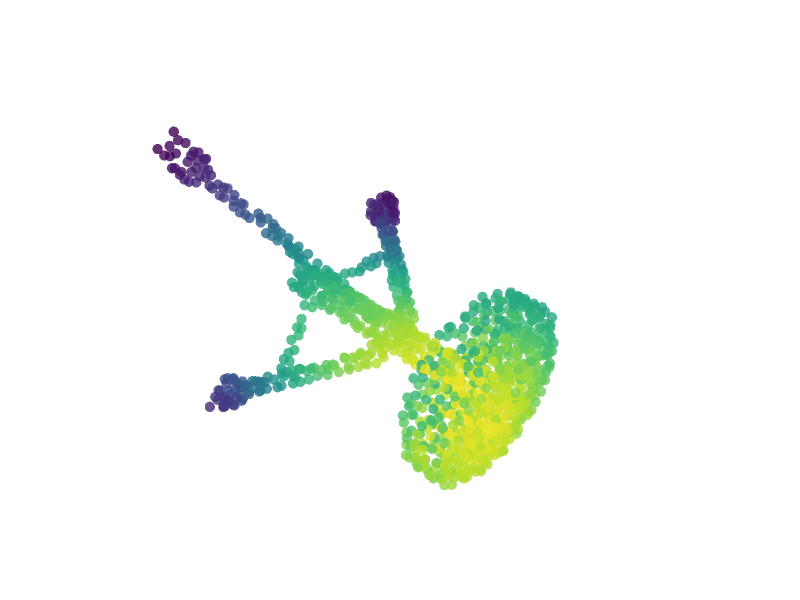} &
      \includegraphics[width=0.11\textwidth]{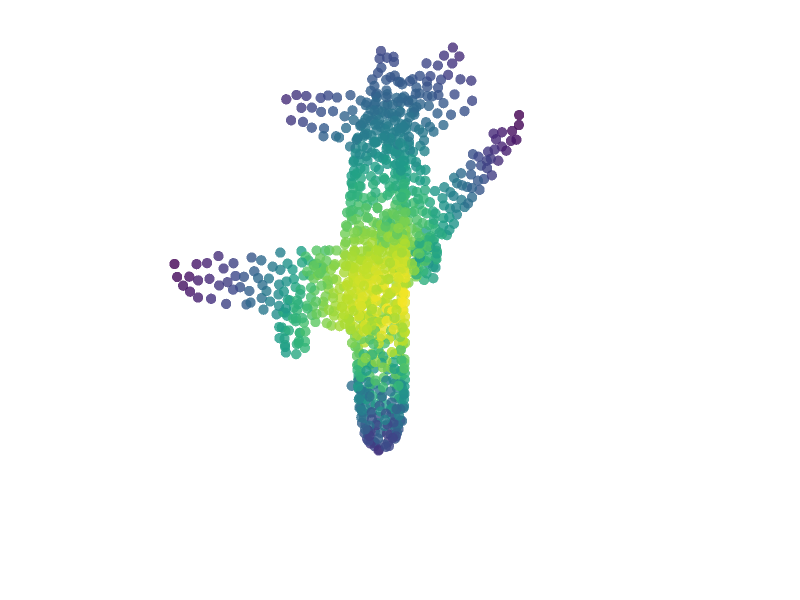} &
      \includegraphics[width=0.11\textwidth]{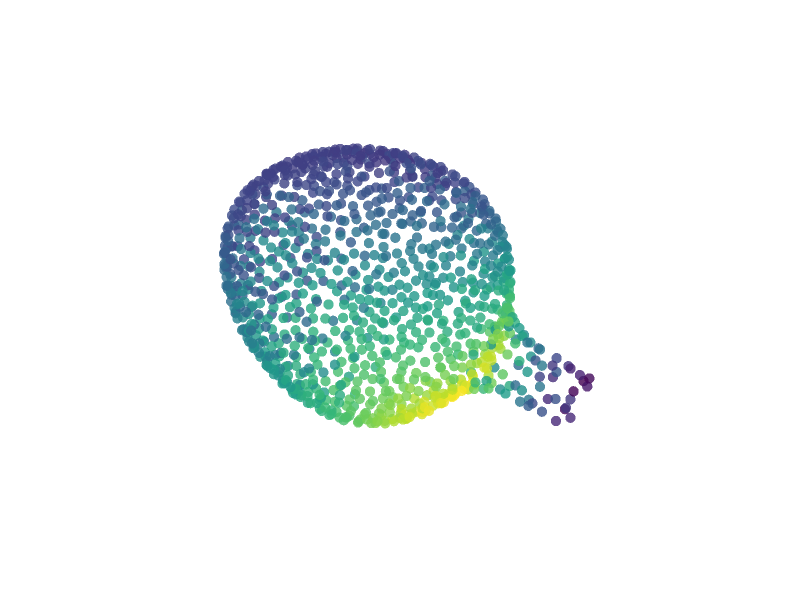} &
      \includegraphics[width=0.11\textwidth]{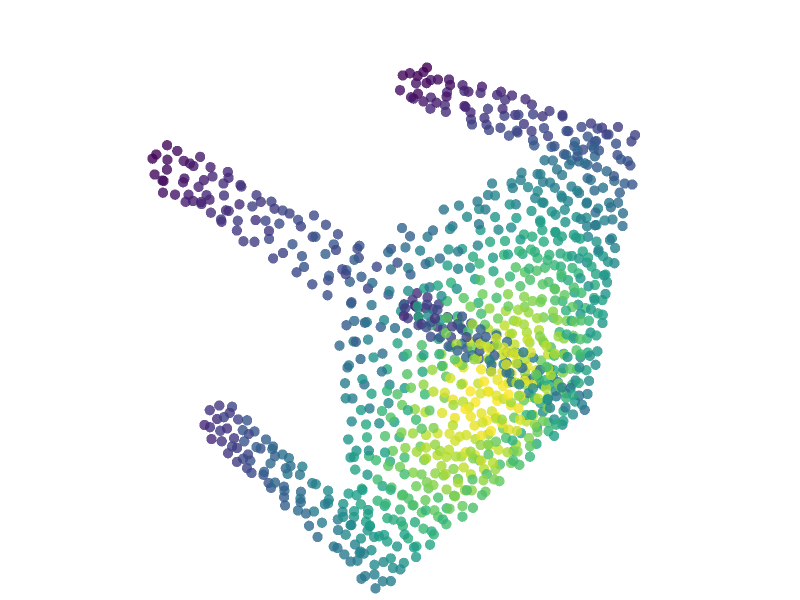} &
      \includegraphics[width=0.11\textwidth]{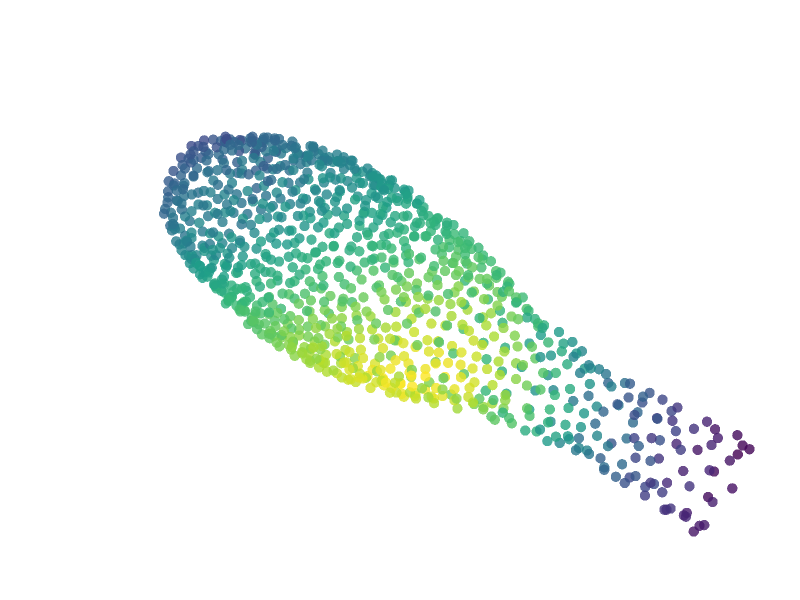} &
      \includegraphics[width=0.11\textwidth]{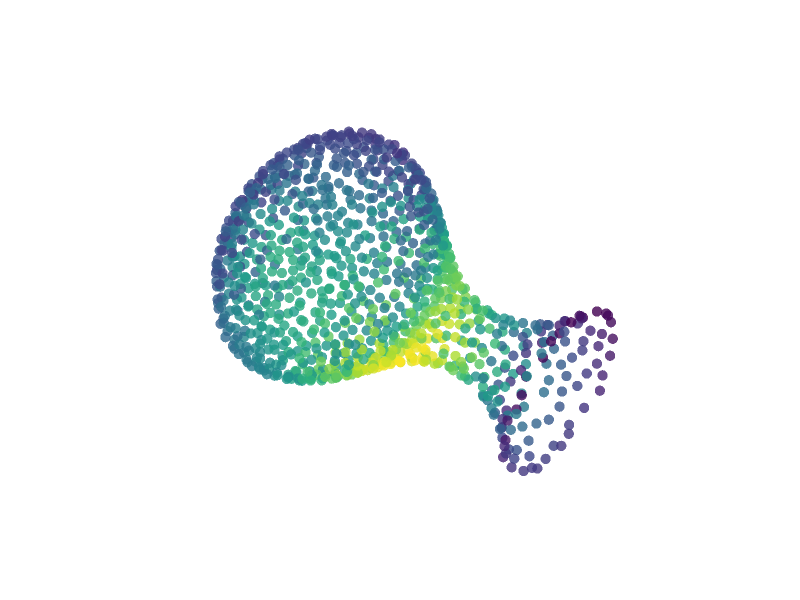} &
      \includegraphics[width=0.11\textwidth]{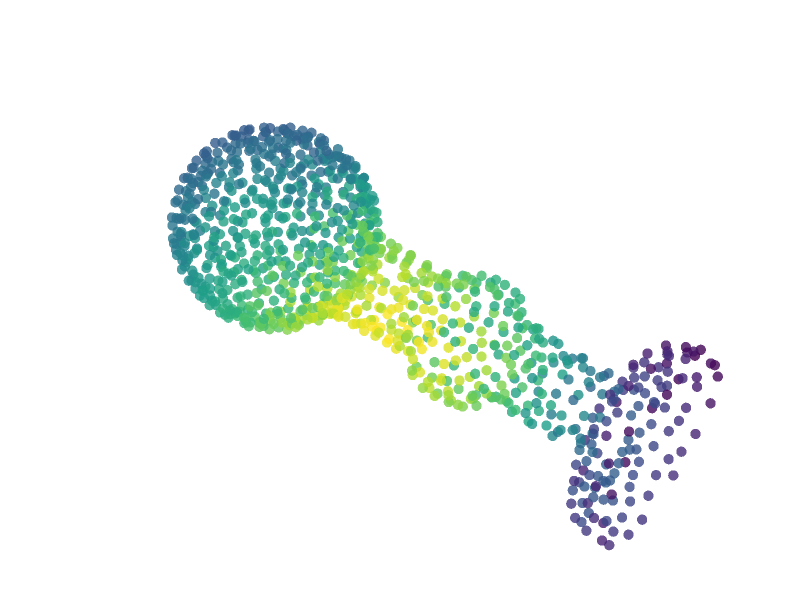} \\

      \raisebox{0.4\height}{\rotatebox{90}{\tiny \textbf{Adaptive32}}} &
      \includegraphics[width=0.11\textwidth]{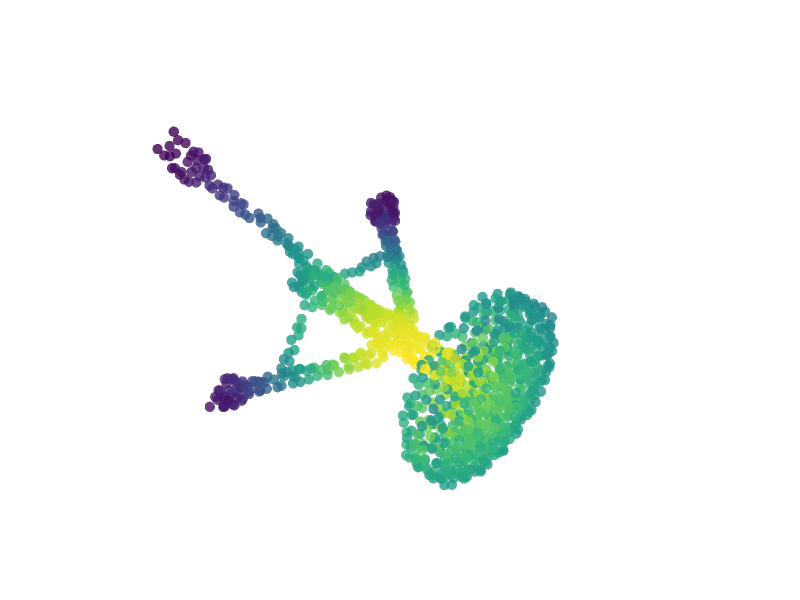} &
      \includegraphics[width=0.11\textwidth]{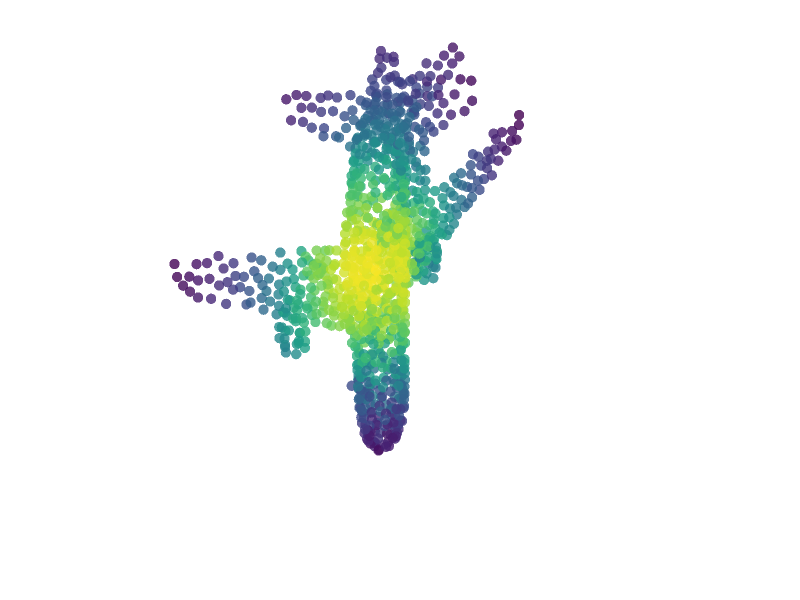} &
      \includegraphics[width=0.11\textwidth]{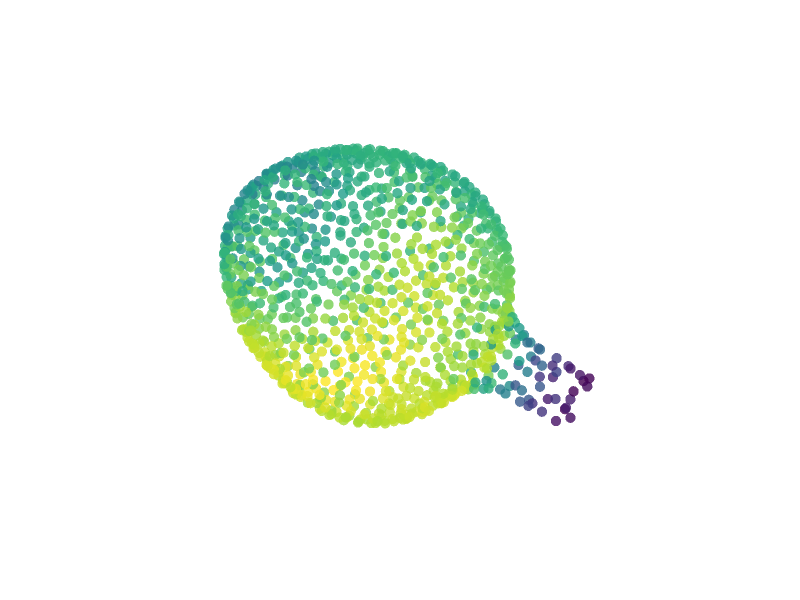} &
      \includegraphics[width=0.11\textwidth]{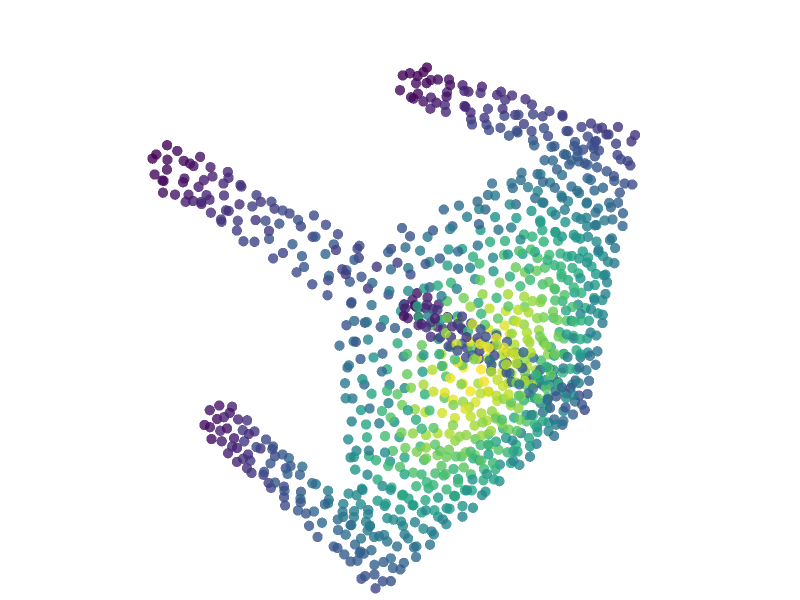} &
      \includegraphics[width=0.11\textwidth]{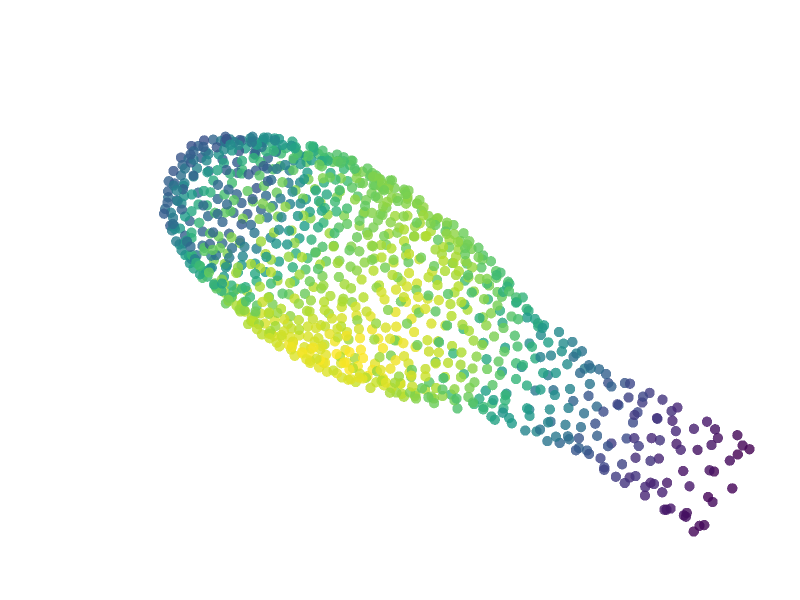} &
      \includegraphics[width=0.11\textwidth]{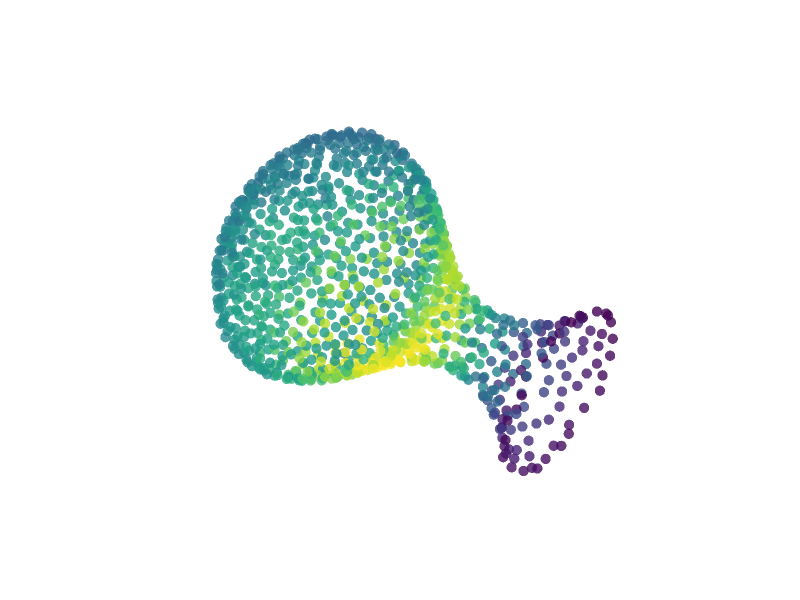} &
      \includegraphics[width=0.11\textwidth]{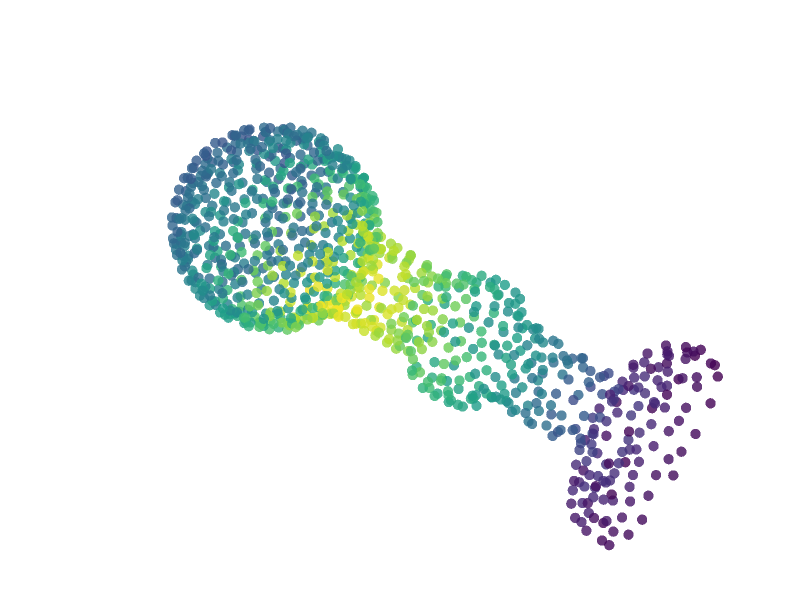} \\

    \end{tabular}%
  }
  \caption{Qualitative attention visualization. Top: DGCNN attention after the second EdgeConv block. Middle/Bottom: activation maps from the proposed AdaptiveEmbedding module with output dimension 16 and 32, respectively.}

  \label{fig:attention_maps}
\end{figure*}

\noindent \textbf{Protocol and reproducibility.}
All experiments use $N{=}1{,}024$ points per shape unless stated. Some parametric baselines in Tab.~\ref{tab:shapenet} report 2k-point results; our efficiency numbers (Tab.~\ref{tab:efficiency}) are measured at 1k points for a consistent runtime/memory comparison. We fix random seeds and evaluate on a single RTX~3090 (24\,GB). Code and scripts to reproduce tables and figures are publicly available.

\subsection{Ablation Studies}
\label{sec:ablation}

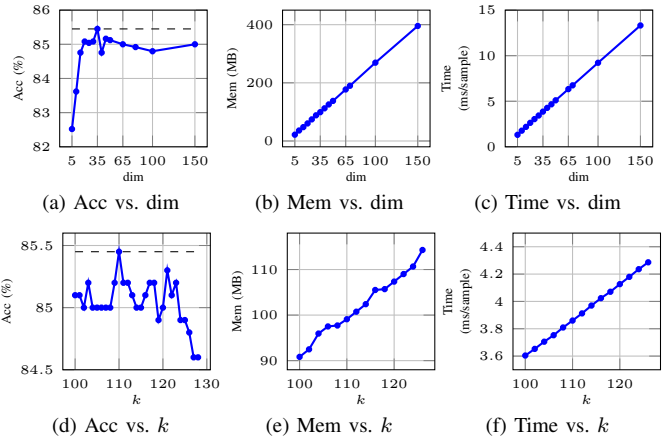
\begin{figure}[t]
  \centering
  \captionsetup[subfigure]{justification=centering, font=footnotesize, skip=1pt}

  \hspace{-0.1\columnwidth}
  \begin{subfigure}[t]{0.4\columnwidth}
    \centering
    \begin{tikzpicture}
      \begin{axis}[
        width=\linewidth,
        height=3.4cm,
        xlabel={dim},
        ylabel={Acc (\%)},
        xlabel near ticks,
        xlabel style={yshift=4pt},
        ylabel near ticks,
        ylabel style={yshift=-4pt},
        ymin=82, ymax=86,
        xtick={5,35,65,100,150},
        grid=major,
        tick label style={font=\tiny},
        label style={font=\tiny},
      ]
      \addplot[mark=*, mark size=1pt, thick, blue] coordinates {
        (5,82.52032399) (10,83.61788392) (15,84.75609422) (20,85.08129716)
        (25,85.04064679) (30,85.08129716) (35,85.45) (40,84.75609422)
        (45,85.16259789) (50,85.12194753) (65,84.99999642) (80,84.91869569)
        (100,84.79674459) (150,84.99999642)
      };
      \addplot[dashed] coordinates {(5,85.45) (150,85.45)};
      \end{axis}
    \end{tikzpicture}
    \caption{Acc vs.\ dim}
    \label{fig:dim_acc_onecol}
  \end{subfigure}
  \hspace{-0.1\columnwidth}
  \begin{subfigure}[t]{0.4\columnwidth}
    \centering
    \begin{tikzpicture}
      \begin{axis}[
        width=\linewidth,
        height=3.4cm,
        xlabel={dim},
        ylabel={Mem (MB)},
        xlabel near ticks,
        xlabel style={yshift=4pt},
        ylabel near ticks,
        ylabel style={yshift=-4pt},
        ymin=-20, ymax=450,
        xtick={5,35,65,100,150},
        grid=major,
        tick label style={font=\tiny},
        label style={font=\tiny},
      ]
      \addplot[mark=*, mark size=1pt, thick, blue] coordinates {
        (5,21.68701172) (10,36.05175781) (15,47.66796875) (20,60.38867188)
        (25,74.30664063) (30,88.3046875) (35,99.09082031) (40,112.3134766)
        (45,125.6708984) (50,137.7924805) (65,177.2192383) (70,189.8383789)
        (100,269.3203125) (150,395.6074219)
      };
      \end{axis}
    \end{tikzpicture}
    \caption{Mem vs.\ dim}
    \label{fig:dim_mem_onecol}
  \end{subfigure}
  \hspace{-0.1\columnwidth}
  \begin{subfigure}[t]{0.4\columnwidth}
    \centering
    \begin{tikzpicture}
      \begin{axis}[
        width=\linewidth,
        height=3.4cm,
        xlabel={dim},
        ylabel=\shortstack{Time\\(ms/sample)},
        xlabel near ticks,
        xlabel style={yshift=4pt},
        ylabel near ticks,
        ylabel style={yshift=-4pt},
        ymin=0, ymax=15,
        xtick={5,35,65,100,150},
        grid=major,
        tick label style={font=\tiny},
        label style={font=\tiny},
      ]
      \addplot[mark=*, mark size=1pt, thick, blue] coordinates {
        (5,1.31563558) (10,1.769655532) (15,2.187366017) (20,2.630930775)
        (25,3.039592168) (30,3.43485381) (35,3.86) (40,4.272817517)
        (45,4.664025842) (50,5.102567255) (65,6.332513647) (70,6.744177825)
        (100,9.221851068) (150,13.31817568)
      };
      \end{axis}
    \end{tikzpicture}
    \caption{Time vs.\ dim}
    \label{fig:dim_time_onecol}
  \end{subfigure}

  \vspace{0.6em}

  \hspace{-0.1\columnwidth}
  \begin{subfigure}[t]{0.4\columnwidth}
    \centering
    \begin{tikzpicture}
      \begin{axis}[
        width=\linewidth,
        height=3.4cm,
        xlabel={$k$},
        ylabel={Acc (\%)},
        xlabel near ticks,
        xlabel style={yshift=4pt},
        ylabel near ticks,
        ylabel style={yshift=-4pt},
        ymin=84.5, ymax=85.6,
        xtick={100,110,120,130},
        grid=major,
        tick label style={font=\tiny},
        label style={font=\tiny},
      ]
      \addplot[mark=*, mark size=1pt, thick, blue] coordinates {
        (100,85.1) (101,85.1) (102,85.0) (103,85.2) (104,85.0)
        (105,85.0) (106,85.0) (107,85.0) (108,85.0) (109,85.2)
        (110,85.45) (111,85.2) (112,85.2) (113,85.1) (114,85.0)
        (115,85.0) (116,85.1) (117,85.2) (118,85.2) (119,84.9)
        (120,85.0) (121,85.3) (122,85.1) (123,85.2) (124,84.9)
        (125,84.9) (126,84.8) (127,84.6) (128,84.6)
      };
      \addplot[dashed] coordinates {(100,85.45) (128,85.45)};
      \end{axis}
    \end{tikzpicture}
    \caption{Acc vs.\ $k$}
    \label{fig:k_acc_onecol}
  \end{subfigure}
  \hspace{-0.1\columnwidth}
  \begin{subfigure}[t]{0.4\columnwidth}
    \centering
    \begin{tikzpicture}
      \begin{axis}[
        width=\linewidth,
        height=3.4cm,
        xlabel={$k$},
        ylabel={Mem (MB)},
        xlabel near ticks,
        xlabel style={yshift=4pt},
        ylabel near ticks,
        ylabel style={yshift=-4pt},
        ymin=88, ymax=118,
        xtick={100,110,120,130},
        grid=major,
        tick label style={font=\tiny},
        label style={font=\tiny},
      ]
      \addplot[mark=*, mark size=1pt, thick, blue] coordinates {
        (100,90.82910156) (102,92.48144531) (104,95.94628906) (106,97.50683594)
        (108,97.71582031) (110,99.09082031) (112,100.7431641) (114,102.3955078)
        (116,105.5117188) (118,105.7001953) (120,107.3525391) (122,109.0048828)
        (124,110.6572266) (126,114.2900391)
      };
      \end{axis}
    \end{tikzpicture}
    \caption{Mem vs.\ $k$}
    \label{fig:k_mem_onecol}
  \end{subfigure}
  \hspace{-0.1\columnwidth}
  \begin{subfigure}[t]{0.4\columnwidth}
    \centering
    \begin{tikzpicture}
      \begin{axis}[
        width=\linewidth,
        height=3.4cm,
        xlabel={$k$},
        ylabel=\shortstack{Time\\(ms/sample)},
        xlabel near ticks,
        xlabel style={yshift=4pt},
        ylabel near ticks,
        ylabel style={yshift=-4pt},
        ymin=3.5, ymax=4.5,
        xtick={100,110,120,130},
        grid=major,
        tick label style={font=\tiny},
        label style={font=\tiny},
      ]
      \addplot[mark=*, mark size=1pt, thick, blue] coordinates {
        (100,3.60403008) (102,3.652663932) (104,3.705299632) (106,3.7527615)
        (108,3.809315668) (110,3.86) (112,3.912385171) (114,3.969812284)
        (116,4.023691112) (118,4.070693455) (120,4.126384594) (122,4.17968286)
        (124,4.235812921) (126,4.286247272)
      };
      \end{axis}
    \end{tikzpicture}
    \caption{Time vs.\ $k$}
    \label{fig:k_time_onecol}
  \end{subfigure}
  \caption{ModelNet40 ablation (classification). Top: varying embedding dimension $d$. Bottom: varying neighborhood size $k$. Columns show accuracy, GPU memory, and inference time.}
  
  \label{fig:ablation_dim_k}
\end{figure}

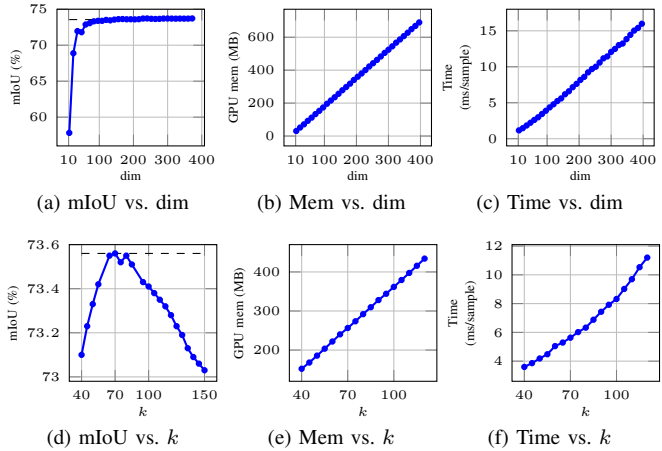
\begin{figure}[t]
  \centering
  \captionsetup[subfigure]{justification=centering, font=footnotesize, skip=1pt}

  \hspace{-0.1\columnwidth}
  \begin{subfigure}[t]{0.4\columnwidth}
    \centering
    \begin{tikzpicture}
      \begin{axis}[
        width=\linewidth,
        height=3.4cm,
        xlabel={dim},
        ylabel={mIoU (\%)},
        xlabel near ticks,
        xlabel style={yshift=4pt},
        ylabel near ticks,
        ylabel style={yshift=-4pt},
        ymin=56, ymax=75,
        xtick={10,100,200,300,400},
        grid=major,
        tick label style={font=\tiny},
        label style={font=\tiny},
      ]
      \addplot[mark=*, mark size=1pt, thick, blue] coordinates {
        (12,57.8122) (24,68.8845) (36,71.9591) (48,71.8088) (60,72.8754)
        (72,73.0668) (84,73.3197) (96,73.3861) (108,73.3876) (120,73.5158)
        (132,73.4556) (144,73.5582) (156,73.6214) (168,73.6357) (180,73.6061)
        (192,73.6070) (204,73.5909) (216,73.6155) (228,73.7070) (240,73.7177)
        (252,73.6723) (264,73.6469) (276,73.6604) (288,73.7051) (300,73.7087)
        (312,73.7013) (324,73.6888) (336,73.7004) (348,73.6923) (360,73.6950) (372,73.72)
      };
      \addplot[dashed] coordinates {(12,73.56) (360,73.56)};
      \end{axis}
    \end{tikzpicture}
    \caption{mIoU vs.\ dim}
    \label{fig:shapenet_dim_acc_onecol}
  \end{subfigure}
  \hspace{-0.1\columnwidth}
  \begin{subfigure}[t]{0.4\columnwidth}
    \centering
    \begin{tikzpicture}
      \begin{axis}[
        width=\linewidth,
        height=3.4cm,
        xlabel={dim},
        ylabel={GPU mem (MB)},
        xlabel near ticks,
        xlabel style={yshift=4pt},
        ylabel near ticks,
        ylabel style={yshift=-4pt},
        ymin=-60, ymax=770,
        xtick={10,100,200,300,400},
        grid=major,
        tick label style={font=\tiny},
        label style={font=\tiny},
      ]
      \addplot[mark=*, mark size=1pt, thick, blue] coordinates {
        (12,30.65625) (24,51.5625) (36,71.06445313) (48,92.28320313) (60,111.6894531)
        (72,133.2558594) (84,153.0683594) (96,173.5683594) (108,196.0703125) (120,215.5703125)
        (132,235.4492188) (144,256.3867188) (156,278.0742188) (168,297.9667969) (180,317.9511719)
        (192,340.5761719) (204,360.0800781) (216,379.8300781) (228,400.4570313) (240,422.5820313)
        (252,442.0820313) (264,462.3359375) (276,482.9609375) (288,504.8984375) (300,524.2128906)
        (312,544.8378906) (324,567.0917969) (336,586.5917969) (348,606.7167969) (360,627.34375)
        (372,649.09375) (384,668.59375) (396,690.0048828125)
      };
      \end{axis}
    \end{tikzpicture}
    \caption{Mem vs.\ dim}
    \label{fig:shapenet_dim_mem_onecol}
  \end{subfigure}
  \hspace{-0.1\columnwidth}
  \begin{subfigure}[t]{0.4\columnwidth}
    \centering
    \begin{tikzpicture}
      \begin{axis}[
        width=\linewidth,
        height=3.4cm,
        xlabel={dim},
        ylabel=\shortstack{Time\\(ms/sample)},
        xlabel near ticks,
        xlabel style={yshift=4pt},
        ylabel near ticks,
        ylabel style={yshift=-4pt},
        ymin=-1, ymax=18,
        xtick={10,100,200,300,400},
        grid=major,
        tick label style={font=\tiny},
        label style={font=\tiny},
      ]
      \addplot[mark=*, mark size=1pt, thick, blue] coordinates {
        (12,1.156359421) (24,1.464776902) (36,1.837722647) (48,2.222357629) (60,2.61209035)
        (72,2.983328725) (84,3.436761344) (96,3.883352536) (108,4.3537853) (120,4.7826506)
        (132,5.238435158) (144,5.605139324) (156,6.180160114) (168,6.596861187) (180,7.160676238)
        (192,7.628095865) (204,8.173773247) (216,8.598798154) (228,9.222939911) (240,9.682376489)
        (252,10.00452622) (264,10.58271287) (276,11.17446758) (288,11.44286163) (300,12.07290669)
        (312,12.48410259) (324,13.01527305) (336,13.23120496) (348,13.84400071) (360,14.43629787)
        (372,15.03003356) (384,15.4069290844223) (396,15.9838952023396)
      };
      \end{axis}
    \end{tikzpicture}
    \caption{Time vs.\ dim}
    \label{fig:shapenet_dim_time_onecol}
  \end{subfigure}

  \vspace{0.6em}

  \hspace{-0.1\columnwidth}
  \begin{subfigure}[t]{0.4\columnwidth}
    \centering
    \begin{tikzpicture}
      \begin{axis}[
        width=\linewidth,
        height=3.4cm,
        xlabel={$k$},
        ylabel={mIoU (\%)},
        xlabel near ticks,
        xlabel style={yshift=4pt},
        ylabel near ticks,
        ylabel style={yshift=-4pt},
        ymin=72.98, ymax=73.6,
        xtick={40,70,100,150},
        grid=major,
        tick label style={font=\tiny},
        label style={font=\tiny},
      ]
      \addplot[mark=*, mark size=1pt, thick, blue] coordinates {
        (40,73.10) (45,73.23) (50,73.33) (55,73.42) (65,73.55)
        (70,73.56) (75,73.52) (80,73.55) (85,73.51) (95,73.43)
        (100,73.41) (105,73.38) (110,73.35) (115,73.32) (120,73.28)
        (125,73.23) (130,73.19) (135,73.13) (140,73.09) (145,73.06) (150,73.03)
      };
      \addplot[dashed] coordinates {(40,73.56) (150,73.56)};
      \end{axis}
    \end{tikzpicture}
    \caption{mIoU vs.\ $k$}
    \label{fig:shapenet_k_acc_onecol}
  \end{subfigure}
  \hspace{-0.1\columnwidth}
  \begin{subfigure}[t]{0.4\columnwidth}
    \centering
    \begin{tikzpicture}
      \begin{axis}[
        width=\linewidth,
        height=3.4cm,
        xlabel={$k$},
        ylabel={GPU mem (MB)},
        xlabel near ticks,
        xlabel style={yshift=4pt},
        ylabel near ticks,
        ylabel style={yshift=-4pt},
        ymin=120, ymax=470,
        xtick={40,70,100,150},
        grid=major,
        tick label style={font=\tiny},
        label style={font=\tiny},
      ]
      \addplot[mark=*, mark size=1pt, thick, blue] coordinates {
        (40,152.0546875) (45,168.0371094) (50,185.6445313) (55,203.5019531)
        (60,221.859375) (65,240.2167969) (70,256.3867188) (75,273.6816406)
        (80,291.9140625) (85,309.6464844) (90,328.0039063) (95,344.1113281)
        (100,361.71875) (105,379.3261719) (110,397.4335938) (115,415.7910156)
        (120,434.1484375)
      };
      \end{axis}
    \end{tikzpicture}
    \caption{Mem vs.\ $k$}
    \label{fig:shapenet_k_mem_onecol}
  \end{subfigure}
  \hspace{-0.1\columnwidth}
  \begin{subfigure}[t]{0.4\columnwidth}
    \centering
    \begin{tikzpicture}
      \begin{axis}[
        width=\linewidth,
        height=3.4cm,
        xlabel={$k$},
        ylabel=\shortstack{Time\\(ms/sample)},
        xlabel near ticks,
        xlabel style={yshift=4pt},
        ylabel near ticks,
        ylabel style={yshift=-4pt},
        ymin=2.6, ymax=12.1,
        xtick={40,70,100,130,150},
        grid=major,
        tick label style={font=\tiny},
        label style={font=\tiny},
      ]
      \addplot[mark=*, mark size=1pt, thick, blue] coordinates {
        (40,3.598160668) (45,3.861431411) (50,4.185072411) (55,4.480100994)
        (60,5.045616523) (65,5.291559392) (70,5.63) (75,6.01900698)
        (80,6.329504251) (85,6.878945673) (90,7.431620406) (95,7.917530783)
        (100,8.321140774) (105,9.01867501) (110,9.695781354) (115,10.53045218)
        (120,11.19779107)
      };
      \end{axis}
    \end{tikzpicture}
    \caption{Time vs.\ $k$}
    \label{fig:shapenet_k_time_onecol}
  \end{subfigure}
  \caption{ShapeNetPart ablation (segmentation). Top row: varying embedding dimension $d$. Bottom row: varying neighborhood size $k$. Columns show instance mIoU, GPU memory, and inference time.}

  \label{fig:shapenet_dim_k}
\end{figure}

\begin{figure}[t]
  \centering
  \captionsetup[subfigure]{justification=centering, font=footnotesize, skip=1pt}

  \begin{subfigure}[t]{0.49\columnwidth}
    \centering
    \begin{tikzpicture}
      \begin{axis}[
        width=\linewidth,
        height=3.4cm,
        xlabel={Stages},
        ylabel={Accuracy (\%)},
        xlabel near ticks,
        xlabel style={yshift=4pt},
        ylabel near ticks,
        ylabel style={yshift=-4pt},
        ymin=82, ymax=86,
        xtick={2,3,4},
        grid=major,
        tick label style={font=\tiny},
        label style={font=\tiny},
      ]
      \addplot[mark=*, mark size=1pt, thick, blue] coordinates {
        (2,83.1) (3,84.7) (4,85.45)
      };
      \addplot[dashed] coordinates {(2,85.45) (4,85.45)};
      \end{axis}
    \end{tikzpicture}
    \caption{ModelNet40 accuracy}
    \label{fig:stages_mn40_onecol}
  \end{subfigure}
  \hfill
  \begin{subfigure}[t]{0.49\columnwidth}
    \centering
    \begin{tikzpicture}
      \begin{axis}[
        width=\linewidth,
        height=3.4cm,
        xlabel={Stages},
        ylabel={mIoU (\%)},
        xlabel near ticks,
        xlabel style={yshift=4pt},
        ylabel near ticks,
        ylabel style={yshift=-4pt},
        ymin=70, ymax=74,
        xtick={1,2,3,4},
        grid=major,
        tick label style={font=\tiny},
        label style={font=\tiny},
      ]
      \addplot[mark=*, mark size=1pt, thick, blue] coordinates {
        (1,73.27) (2,73.56) (3,72.47) (4,70.90)
      };
      \addplot[dashed] coordinates {(1,73.56) (4,73.56)};
      \end{axis}
    \end{tikzpicture}
    \caption{ShapeNetPart mIoU}
    \label{fig:stages_shapenet_onecol}
  \end{subfigure}
  \caption{Effect of stage depth. (a) ModelNet40 classification accuracy vs.\ number of stages. (b) ShapeNetPart instance mIoU vs.\ number of stages.}

  \label{fig:stages_combined}
  \vspace{-10pt} 
\end{figure}
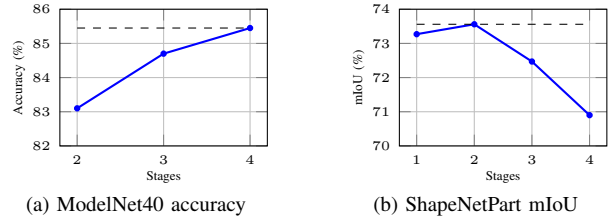

We ablate the key claim: \emph{input-aware adaptivity} in the positional encoding is necessary for stable, training-free performance across datasets and tasks. We use ModelNet40 for classification and ShapeNetPart for segmentation.
Each sweep varies one factor while holding others fixed: (i) adaptivity (Fig.~\ref{fig:fix}); (ii) embedding dimension $d$ (Figs.~\ref{fig:ablation_dim_k},~\ref{fig:shapenet_dim_k}); (iii) neighborhood size $k$ (Figs.~\ref{fig:ablation_dim_k},~\ref{fig:shapenet_dim_k}); and (iv) stage depth (Fig.~\ref{fig:stages_combined}).

\noindent \textbf{Adaptivity of bandwidth and blend.}
Fixing $\sigma$ or the Gaussian/cosine blend consistently hurts performance (Figs.~\ref{fig:fixsigma},~\ref{fig:fixblend}). Accuracy peaks in a narrow window and drops quickly outside it, while the adaptive scheme maintains strong results without per-dataset tuning. This supports the use of per-input $\sigma$ and $\lambda$.

\noindent \textbf{Embedding dimension.}
Classification saturates around $d=30$--40 (Fig.~\ref{fig:dim_acc_onecol}); we use $d=35$ to stay near the peak while keeping cost low.
In contrast, segmentation benefits from larger embeddings and saturates later (Fig.~\ref{fig:shapenet_dim_acc_onecol}); we choose $d=144$ as a balanced point before the marginal mIoU gains become small relative to the additional compute and memory.
This task-dependent behavior is expected: classification mainly requires a global shape signature, whereas segmentation must preserve part-level cues and boundary detail, which benefit from higher-capacity embeddings.
As $d$ increases, GPU memory and inference time grow approximately linearly for both tasks (Figs.~\ref{fig:dim_mem_onecol}--\ref{fig:dim_time_onecol} and \ref{fig:shapenet_dim_mem_onecol}--\ref{fig:shapenet_dim_time_onecol}), making $d$ an effective knob for trading accuracy against efficiency.

\noindent \textbf{Neighborhood size $k$.}
Accuracy peaks near $k=110$ on ModelNet40 (Fig.~\ref{fig:k_acc_onecol}) and near $k=70$ on ShapeNetPart (Fig.~\ref{fig:shapenet_k_acc_onecol}).
Larger neighborhoods provide more geometric context, which can help classification up to a point; however, overly large $k$ can dilute local detail that is important for part boundaries and can interact unfavorably with downsampling in deeper hierarchies.
Consistent with this intuition, ShapeNetPart prefers a moderate $k$, while ModelNet40 tolerates a larger neighborhood for stronger context aggregation.
In all cases, increasing $k$ raises GPU memory and inference time steadily because neighborhood grouping dominates the encoder cost
(Figs.~\ref{fig:k_mem_onecol}--\ref{fig:k_time_onecol} and \ref{fig:shapenet_k_mem_onecol}--\ref{fig:shapenet_k_time_onecol}), so $k$ is a direct lever for reducing latency when needed.

\noindent \textbf{Number of stages.}
Using more stages improves classification, with 4 stages performing best on ModelNet40 (Fig.~\ref{fig:stages_combined}).
Deeper hierarchies increase receptive field and multi-scale aggregation, which benefits global recognition.
For segmentation, using more than 2 stages reduces mIoU and increases cost (Fig.~\ref{fig:stages_combined}), likely because aggressive downsampling and broad aggregation blur fine-grained part boundaries.
Accordingly, we use 4 stages for classification and 2 stages for segmentation as a simple, robust default.

\noindent \textbf{Resource--accuracy trade-offs and practical tuning.}
Across both tasks, the ablations reveal a trade-off: accuracy (or mIoU) improves with larger $d$ and richer neighborhoods, while GPU memory and inference time grow nearly monotonically with $d$ and $k$ (Figs.~\ref{fig:ablation_dim_k}--\ref{fig:shapenet_dim_k}). In practice, $d$ is a coarse knob that sets representational capacity, whereas $k$ controls the locality--context balance. For ModelNet40, performance saturates quickly with $d$ and exhibits a clear peak with $k$, so a moderate embedding with a moderately large neighborhood is sufficient. In contrast, ShapeNetPart benefits from larger embeddings to capture part-level structure, but overly large neighborhoods and deeper hierarchies can hurt because aggressive downsampling and broad aggregation blur fine boundaries.
hu
From a deployment perspective, these trends provide simple guidelines. If memory is the bottleneck, reducing $d$ yields the most direct savings with limited accuracy loss near the saturation region. If inference time is the bottleneck, reducing $k$ and/or stage depth offers the largest speedups because neighborhood aggregation dominates runtime. Finally, the adaptivity in $\sigma$ and the Gaussian--cosine mixing reduces the need to retune these hyperparameters across datasets with different sampling densities (Fig.~\ref{fig:fix}), allowing $d$, $k$, and stages to serve as the main task-dependent controls.

\noindent \textbf{Qualitative comparison.}
Fig.~\ref{fig:attention_maps} compares attention maps from the parametric DGCNN with those produced by the proposed AdaptiveEmbedding module. Despite being non-parametric, the adaptive embedding yields more coherent and part-aware responses, indicating that it captures meaningful geometric structure without learned weights.

\subsection{Takeaways}
NPNet demonstrates that well-designed non-parametric components---particularly the adaptive Gaussian--Fourier positional encoding---can substantially close the gap to parametric networks on both classification and segmentation. The method is especially appealing in few-shot and rapid-deployment scenarios because it eliminates the training step and adapts to the input geometry at inference time. Ablations provide clear guidance for selecting $d$, $k$, and stage depth to meet differing accuracy and efficiency requirements.

\section{Conclusion}
\label{sec:conclusion}

We presented \textbf{NPNet}, a fully non-parametric approach for 3D point-cloud classification and part segmentation. 
Its core is an \emph{adaptive Gaussian--Fourier positional encoding} that adjusts to the input geometry, improving stability across scales and sampling densities without introducing trainable parameters. 
Across ModelNet40, ModelNet-R, ScanObjectNN, and ShapeNetPart, NPNet outperforms prior non-parametric baselines and remains competitive with parametric networks while eliminating training altogether.
NPNet is also efficient in practice: memory and latency are primarily controlled by the embedding dimension and neighborhood size, which makes deployment tuning straightforward. 
Few-shot results further show that NPNet adapts well to limited supervision through prototype-based inference, with no fine-tuning.

Future work will extend NPNet to detection and scene-level tasks and investigate lightweight hybrid designs that preserve the training-free core while adding minimal learnable capacity when needed.





	\bibliographystyle{IEEEtran}
	\bibliography{ref} 
	
\end{document}